\documentclass{bmvc2k}
\usepackage{adjustbox}
\usepackage{multirow}

\title{3D Human Pose Estimation with \\ Relational Networks}

\addauthor{Sungheon Park}{sungheonpark@snu.ac.kr}{1}
\addauthor{Nojun Kwak}{nojunk@snu.ac.kr}{1}

\addinstitution{
 Department of Transdisciplinary Studies\\
 Seoul National University\\
 Seoul, Korea
}

\runninghead{S. Park and N. Kwak}{3D HPE with Relational Networks}

\def\etal{\emph{et al}\bmvaOneDot}

\newcommand\nj[1]{\textcolor{black}{#1}}
\newcommand\sh[1]{\textcolor{black}{#1}}

\begin{document}

\maketitle

\begin{abstract}
In this paper, we propose a novel 3D human pose estimation algorithm from a single image based on neural networks. We adopted the structure of the relational networks in order to capture the relations \nj{among} different body parts. \nj{In our method, each} pair of different body parts generates features, and \nj{the} average of the features \nj{from all the} pairs are used for 3D pose estimation. In addition, we propose a dropout method that can be used in relational modules, which inherently \nj{imposes} robustness to the occlusions. The proposed network achieves state-of-the-art performance for 3D pose estimation in Human 3.6M dataset, and it effectively produces \nj{plausible results} \nj{even in the existence of missing joints}.
\end{abstract}


\section{Introduction}
\label{sec:intro}

Human pose estimation (HPE) is a fundamental task in computer vision, which can be adopted to many applications such as action recognition, human behavior analysis, virtual reality and so on. Estimating 3D pose of human body joints from \nj{2D joint locations} is an under-constrained problem. However, since human joints are connected by rigid bodies, \nj{the} search space of 3D pose is limited to the range of joints. Therefore, it is able to learn 3D structures from 2D positions, and numerous \nj{studies} on 2D-to-3D mapping of human body \sh{have} been conducted. Recently, \nj{\citet{Martinez_2017_ICCV}} proved that a simple fully connected neural network that accepts raw 2D positions as an input gives surprisingly accurate results. Inspired by this result, we designed \nj{a} network that accepts 2D positions of joints as inputs and generates 3D positions \nj{based on them}.

Human body can be divided into arms, legs, \nj{a} head, and \nj{a} torso\nj{, each of which} has distinctive behaviors and movements. We designed the network so that it learns the relations \nj{among} different body parts. The relational modules for the neural networks proposed in \cite{santoro2017} provided a way to learn relations between the components \nj{within a neural network architecture}. We adopt \nj{this} relational modules for 3D \sh{HPE} with a little \nj{modification}. Specifically, the body joints are divided into several groups, and the relations between them are learned via relational networks. The features \sh{from} all pairs of groups are averaged to generate \sh{the feature vectors which are} used for 3D pose regression. We found this simple structure outperforms \nj{the baseline which uses a fully connected network. }
Moreover, we propose a method that can impose robustness to the missing points during the training. The proposed method, named as relational dropout, randomly drops one of the pair features when they are averaged, which simulates the case that \nj{certain} groups of joints are missing during the training.
\nj{To capture the relations among joints within a group, we also designed a hierarchical relational network, which further allows robustness to wrong 2D joint inputs.}
Lastly, we discovered \nj{that the proposed} structure of the network modified from~\cite{Martinez_2017_ICCV} and \nj{the} finetuning schemes \nj{improve} the performance \nj{of HPE}. The proposed method achieved state-of-the-art performance in 3D \sh{HPE} on Human 3.6M dataset~\cite{ionescu2014human3}, and the network \nj{can robustly estimate} 3D poses even when multiple \nj{joints} are missing \nj{using the proposed relational dropout scheme}.

The rest of the paper is organized as follows. \nj{Some of the related} works are introduced in Section~\ref{sec:rel} \nj{and} the proposed method is explained in Section~\ref{sec:method}. \nj{Section~\ref{sec:exp} shows experimental results and finally, Section~\ref{sec:concl} concludes the paper.} 

\section{Related Work}
\label{sec:rel}

Early stage \nj{studies on} 3D HPE from RGB images used hand-crafted \nj{features} such as local shape context~\cite{agarwal2006recovering}, histogram of gradients~\cite{rogez2008randomized,onishi20083d}, or segmentation results~\cite{ionescu2011latent}. \sh{From those features, 3D poses} \nj{were} retrieved via regression using relevance vector machine~\cite{agarwal2006recovering}, randomized trees~\cite{rogez2008randomized}, structured SVMs~\cite{ionescu2011latent}, KD-trees~\cite{Yasin_2016_CVPR} or Bayesian non-parametric models~\cite{sanzari2016bayesian}.

Recent advancements in neural networks boosted the performance of 3D HPE. \nj{\citet{li20143d}} firstly applied convolutional neural networks (CNNs) to 3D HPE. Since then, various models that capture structured representation of human bodies have been combined with CNNs using, for instance, denoising autoencoders~\cite{tekin_2016_BMVC}, maximum-margin cost function~\cite{li2015maximum}, and pose priors from 3D data~\cite{bogo2016keep,lassner2017unite,rogez2017lcr}.

It has been proven that 2D pose information \nj{acts a} crucial role for 3D pose estimation. \nj{\citet{park_eccvw_2016}} directly \nj{propagated} 2D pose estimation results to \nj{the} 3D pose estimation part in a single CNN. \citet{pavlakos2017coarse} proposed a volumetric representation that gradually \nj{increases} the resolution of the depth from \nj{heatmaps of 2D pose}. \citet{mehta2017vnect} similarly \nj{regressed} the position of each coordinate using heatmaps. There are a couple of works that directly regress 3D pose from an \nj{image} using constraints \nj{on} human joints~\cite{sun2017compositional} or combining \nj{weakly}-supervised learning~\cite{zhou2017towards}. \citet{tome2017lifting} lifted 2D pose heatmaps to 3D pose via probabilistic pose models. \citet{Tekin_2017_ICCV} combined features from \nj{both RGB} images and 2D pose heatmaps which were used for 3D pose estimation.

While 3D pose estimation from images \nj{have shown} impressive performance, there is another approach that infers \nj{a} 3D pose directly from \nj{the result of 2D pose estimation}. It usually has a two-stage procedure\nj{: 1)} 2D pose estimation using CNNs and \nj{2)} 3D pose inference via neural networks using the \nj{estimated 2D pose}. \citet{Chen_2017_CVPR} found that \nj{a} non-parametric nearest neighbor model that \nj{estimates a} 3D pose from \nj{a} 2D pose showed comparable performance when \nj{the} precise 2D pose \nj{information is} provided. \nj{\citet{moreno20173d}} proposed a neural network that outputs 3D Euclidean distance matrices from 2D inputs. \citet{Martinez_2017_ICCV} proposed a simple neural network that directly \nj{regresses a} 3D pose from raw 2D joint positions. The network consists of two residual modules~\cite{he2016deep} with batch normalization~\cite{ioffe15} and dropout~\cite{srivastava2014dropout}. The method showed state-of-the-art performance despite its simple structure. The performance has been further improved by recent works. \citet{fang2017learning} proposed a pose grammar network that incorporates a set of knowledge learned from human body\nj{, which was designed} as a bidirectional recurrent neural \nj{network}. \citet{yang20183d} used adversarial learning to implicitly learn geometric \nj{configuration} of human body. \citet{cha2018pose} developed a consensus algorithm that generates \nj{a} 3D pose from multiple partial hypotheses which \nj{are} based on a non-rigid structure from motion algorithm~\cite{lee2016consensus}. The method is similar to our method in that they divided the body joints into multiple groups. However, \nj{our} proposed method integrates the features of all groups within the network rather than generating \nj{a} 3D pose from each group \nj{as in \citet{cha2018pose}}.

There are a few approaches that exploit temporal information using \nj{various methods such as} overcomplete dictionaries~\cite{zhou2016sparseness,zhou2018monocap}, 3D CNNs~\cite{grinciunaite2016human}, sequence-to-sequence networks~\cite{hossain2017exploiting}, \nj{and} multiple-view settings~\cite{Pavlakos_2017_CVPR}. In this paper, we focus on the case that both training and testing are conducted on a single image.

\section{Methods}
\label{sec:method}

\begin{figure}
    \centering
    \hfill
    \subfigure[ ]{\label{fig_1_1}
    {\includegraphics[width=0.34\textwidth]{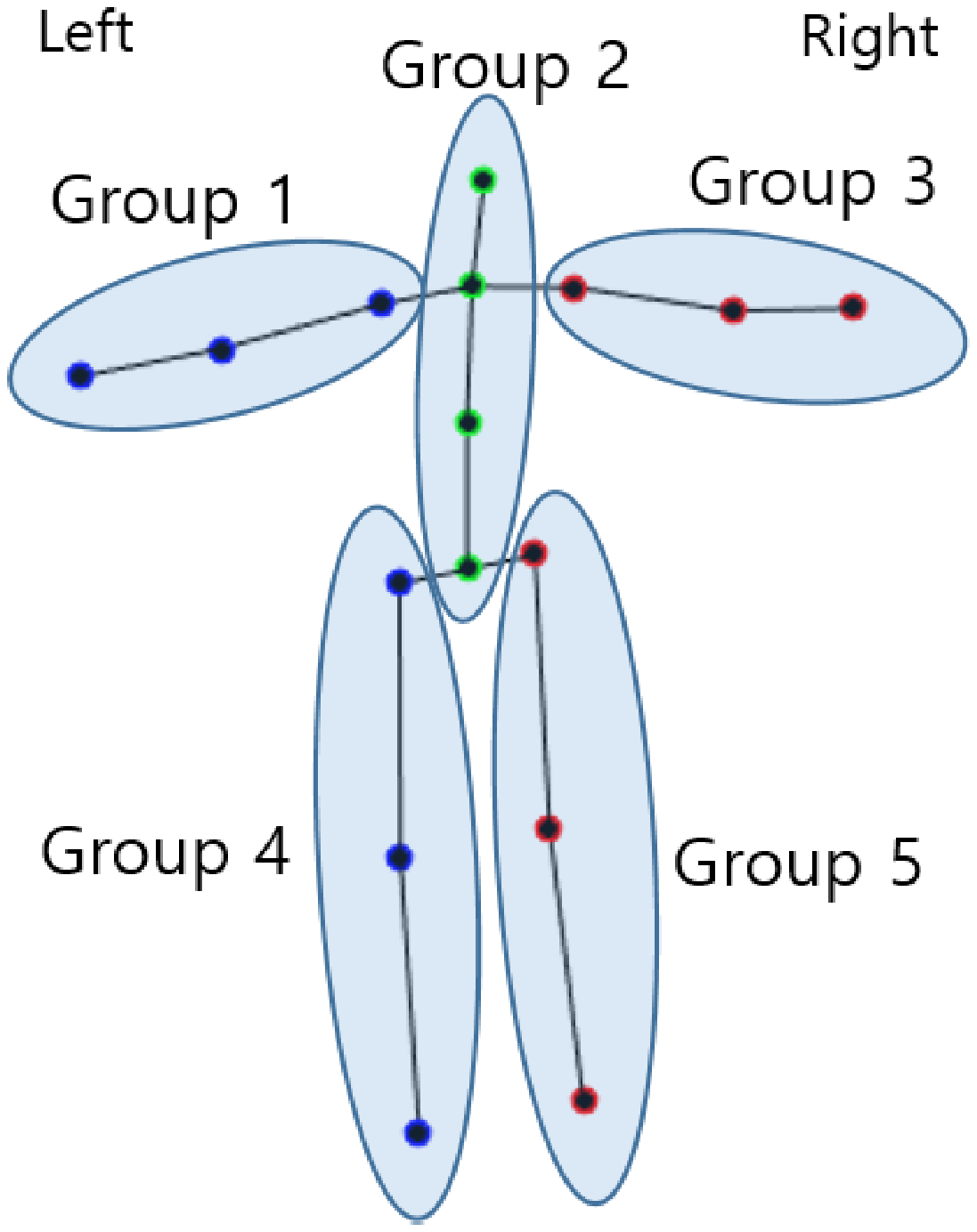}}}\hfill
    \subfigure[ ]{\label{fig_1_2}
    {\includegraphics[width=0.19\textwidth]{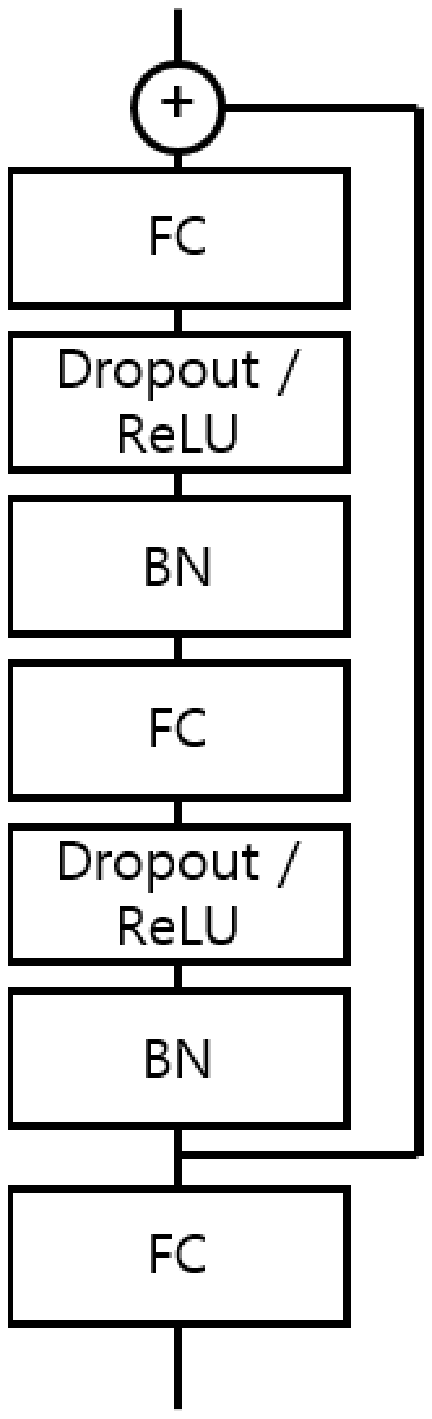}}}\hfill
    \subfigure[ ]{\label{fig_1_3}
    {\includegraphics[width=0.44\textwidth]{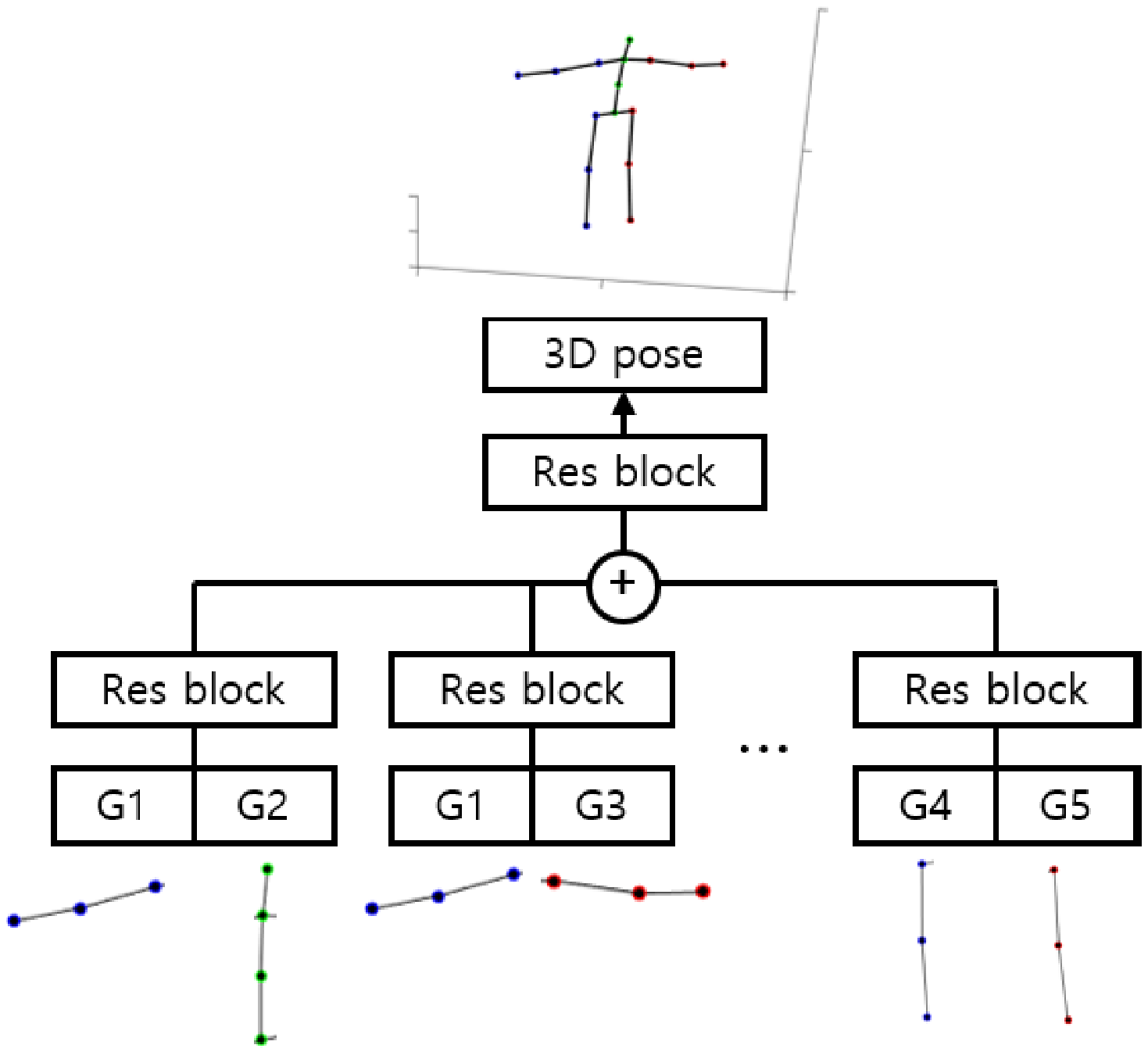}}}\hfill
    
\caption{(a) Group configurations used in this paper. We divided 16 2D input joints to non-overlapping 5 groups each of which corresponds to left/right arms, left/right legs and a torso. (b) The residual module used in this paper. We adopted the structure suggested in~\cite{he2016identity}. (c) The structure of the RN for 3D HPE. Features extracted from all pairs of groups are averaged to produce features for pose estimation. Each Resblock in the figure has the same structure shown in (b).}
\label{fig_1}
\end{figure}

\subsection{Relational Networks for 3D Human Pose Estimation}

Relation networks (RN) proposed in~\cite{santoro2017} consists of two parts, one that does relational reasoning and the other that \nj{performs} a task-specific inference. The output of the RN is formulated as follows:
\begin{equation}
RN(O) = f ( \sum_{(i,j)} g(o_i , o_j) ),
\end{equation}
where $f$ and $g$ are functions that are represented as \nj{corresponding} neural networks, \nj{and $O = \{o_1, \cdots o_n \}$ is the set of objects.} Pairs of different objects \nj{$o_i$, $o_j$} are fed to the network \nj{$g$}, and the relation of all pairs are summed together to generate features that capture relational information.

We adopt the concept and the structure of the RN to 3D human pose estimation. The network proposed in this paper takes $2 n_{2D}$\nj{-dimensional} vectors as inputs and outputs $3 (n_{3D}-1)$\nj{-dimensional} vectors where $n_{2D}$ and $n_{3D}$ are \nj{the number} of 2D and 3D joints respectively. For 2D inputs, we used $(x,y)$ coordinates of detected joints in RGB images whereas relative \nj{positions} of $(x,y,z)$ coordinates from the root joint are estimated for 3D pose estimation. In the \nj{original RN~\cite{santoro2017}}, a neural network module that generates \nj{a pairwise} relation, $g(\cdot)$, shares \nj{weights} across all pairs of objects. This weight sharing makes the network learn order-invariant relations. However, this scheme is not applied to \nj{our} 2D-to-3D regression of human pose \nj{as the following reasons}. While original RN tries to capture the holistic relations that does not depend on the position of the objects or order of pairs, the groups on human body \nj{represent} different parts where order of pairs matters. For instance, if the 2D positions of the left \nj{arm} and \nj{the} right \nj{arm} are switched, \nj{the} 3D pose should also be changed accordingly. However, the relational features generated will be the same for both cases if the order of pair is not considered. For these reasons, we did not use weight sharing for relational models. The 3D HPE algorithm proposed in this paper is formulated as 
\begin{equation}
S_{3D} (S_{2D}) = f ( \frac{1}{n_p} \sum_{(i,j)} g_{i,j} (G_i , G_j) ),
\label{eq2}
\end{equation}
where $n_p$ is the number of pairs, $S_{3D}$, $S_{2D}$ represents 3D and 2D shape of human body joints respectively, and \nj{$G_i$} corresponds to the subsets of 2D input joints \nj{belonging to group $i$}. We divide the input 2D joints to non-overlapping five groups as illustrated in Fig.~\ref{fig_1_1}. \nj{Total} 16 joints are given as an input \nj{to} the proposed network. Each \nj{joint} group contains 3 or 4 joints, which we designed so that each group has \nj{a small range of variations}. Each group represents \nj{a} different part of \nj{a} human body in this configuration. In other words, the groups contain joints from left/right arms, left/right legs, or the rest (\nj{a} head and \nj{a} torso). Thus, the relational network captures how different body parts are related with each other. All pairs of $(i,j)$ such that $i<j$ are fed to the network and generates features of the same dimension. \nj{The mean} of the relational features \nj{is} passed to the next network module that is denoted as $f(\cdot)$ in Eq.~\ref{eq2}. We empirically found that using \nj{the} mean of the relational features instead of \nj{the} sum stabilizes training.

 We used ResNet structures proposed in~\cite{he2016identity} for neural network modules that are used for relation extraction and 3D pose estimation. The structure of a single module is illustrated in Fig.~\ref{fig_1_2}. A fully connected layer is \nj{firstly} applied to increase the input dimension to \nj{that} of \nj{a} feature vector. Then, \nj{a residual network consisting} of two sets of batch normalization~\cite{ioffe15}, dropout~\cite{srivastava2014dropout}, \nj{a} ReLU activation function, and \nj{a} fully connected layer \nj{is} applied. The overall structure of the proposed network for 3D HPE is illustrated in Fig.~\ref{fig_1_3}.

It can be advantageous if we are able to capture the relations of pairs of individual joints. However, in this case, there are total $\frac{n_{2D}(n_{2D}-1)}{2}$ pairs which makes the network quite large. Instead, we designed \nj{a} hierarchical relational network in which relations \nj{between} two joints \nj{in a group} are extracted within the group. The feature of each group $G_k$ is generated as
\begin{equation}
G_k = \frac{1}{n_{p_k}} \sum_{(i,j)} g_{i,j}^k (P_i , P_j),
\label{eq3}
\end{equation}
where $n_{p_k}$ is the number of pairs in group $k$, and $P_i , P_j$ correspond to 2D joints that belong to group $k$. The generated features are used as an input to the next relational network which is formulated as Eq.~\ref{eq2}. Empirically, we observe that the hierarchical representation does not outperform a single level relational network, but the structure \nj{is advantageous if the} relational dropout \nj{is applied as} described in Section~\ref{subsec:rel_drop}.

\subsection{Relational Dropout}
\label{subsec:rel_drop}

In this section, we propose a regularization method, \nj{which we call \textit{`relational dropout'},} that can be applied to relational networks. Similar to dropout~\cite{srivastava2014dropout}, we randomly drop the relational feature vectors that \nj{contain information on} a certain group. In this paper, we restrict the \nj{number of} dropping element \nj{to be} at most 1. Thus, when \nj{the} number of groups is $n_G$, \nj{among the $\frac{n_G (n_G-1)}{2}$ pairs,} $n_G - 1$ relational feature vectors are dropped \nj{and replaced with zero vectors} when relational dropout is applied. After \nj{the} mean of the feature vectors are calculated, it is divided by the portion of non-dropping vectors to maintain the scale of the feature vector as in the \nj{general} dropout \nj{method}. Concretely, when group $k$ is selected to \nj{be dropped}, the formulation becomes
\begin{equation}
S_{3D} (S_{2D} | \mathrm{drop}=k) = f (\frac{1}{n_p - n_G + 1}  \sum_{(i \neq k,j \neq k)} g_{i,j} (G_i , G_j) ).
\label{eq4}
\end{equation}
Dropping features of a certain group simulates the case that the 2D points \nj{belonging} to the dropping group are missing. Hence, the network learns to estimate \nj{the} 3D pose not only when all \nj{the} 2D joints are visible but also when some of them are \nj{invisible}. The relational dropout is applied with the probability of $p_{drop}$ during the training. Since at most one group is dropped, the \nj{combinational variability} of missing joints is limited. To alleviate the problem, we applied \nj{the proposed} relational dropout to hierarchical relational networks. In this case, we are able to simulate the case when a certain joint in a group is missing and \sh{to} simulate various combinations of missing joints. At test time, we simply apply relational dropout to the groups that contain missing points.

\subsection{Implementation Details}
\label{subsec:details}

For the networks used in the experiments, the pose estimator $f(\cdot)$ in the relational networks has fully connected layers of 2,048 dimensions with \nj{a} dropout probability of 0.5. For the modules \nj{$g_{i,j}(\cdot)$} that generates relational feature vector of the pairs $G_i$ and $G_j$, 1,024 dimensional fully connected layers \nj{with a} dropout probability of 0.25 are used. Lastly, for the hierarchical relational networks, the modules that generate relations from the \nj{pairs} of 2D joints consist of 256 dimensional fully connected layers \nj{with a} dropout probability of 0.1. \sh{When the relational dropout is applied during the training, $p_{drop}$ is set to 0.2 for the case that one of the groups of joints is dropped, and it is set to 0.1 when the relational dropout is applied to the hierarchical relational units to drop a single joint}.

We used stacked hourglass network~\cite{newell2016stacked} to infer 2D joint positions from training and test images. We finetuned the network pre-trained on MPII human pose dataset~\cite{andriluka14cvpr} using the frames of Human3.6M dataset. Mean subtraction is the only pre-processing applied to both 2D and 3D joint positions.

The \nj{proposed network is} trained using ADAM optimizer~\cite{kingma2014adam} with \nj{a} starting learning rate of 0.001. \nj{The batch} size is set to 128, and the learning rate \nj{is halved for} every 20,000 iterations. The network is trained for 100,000 iterations.

As a final note, we found that finetuning the trained model to each sequence of Human 3.6M dataset improves the estimation performance. During the finetuning, batch normalization statistics are fixed and the dropout probability is set to 0.5 in all modules.

\section{Experimental Results}
\label{sec:exp}

We used Human 3.6M dataset~\cite{ionescu2014human3} to validate the proposed algorithm. The dataset is the largest dataset for 3D \sh{HPE}, and it consists of 15 action sequences which were performed by 7 different persons. Following the previous works, we used 5 subjects (S1, S5, S6, S7, S8) for training and 2 subjects (S9, S11) for testing. Mean \nj{per-joint} position error (MPJPE) is used \nj{as an} evaluation metric. We reported MPJPE for two types of alignments: aligning the root joints of the estimated pose and the ground truth pose denoted as \textit{Protocol 1}, and aligning via Procrustes analysis including scaling, rotation, and translation denoted as \textit{Protocol 2}. The proposed method is compared to the recently proposed methods that estimates 3D pose from a single image~\cite{pavlakos2017coarse, Martinez_2017_ICCV,fang2017learning,cha2018pose,yang20183d,moreno20173d,zhou2017towards,Tekin_2017_ICCV}.

To compare the performance of the proposed algorithm to the network that does not use relational networks, we designed \nj{a baseline} network \nj{containing only} fully connected layers. The baseline network consists of two consecutive ResBlocks of 2,048 dimensions. Dropout with probability of 0.5 is applied.

\begin{table}[t]
\begin{center}
\adjustbox{width=0.8\linewidth}{
\begin{tabular}{|l|c @{\hskip 0.3em} c @{\hskip 0.3em} c @{\hskip 0.3em} c @{\hskip 0.3em} c @{\hskip 0.3em} c @{\hskip 0.3em} c @{\hskip 0.3em} c|}
\hline
Method & Direct & Discuss & Eat & Greet & Phone & Photo & Pose & Purchase \\
\hline
Pavlakos \etal~\cite{pavlakos2017coarse} & 67.4 & 71.9 & 66.7 & 69.1 & 72.0 & 77.0 & 65.0 & 68.3 \\
Tekin \etal~\cite{Tekin_2017_ICCV} & 54.2 & 61.4 & 60.2 & 61.2 & 79.4 & 78.3 & 63.1 & 81.6 \\
Zhou \etal~\cite{zhou2017towards} & 54.8 & 60.7 & 58.2 & 71.4 & 62.0 & 65.5 & 53.8 & 55.6 \\
Martinez \etal~\cite{Martinez_2017_ICCV} & 51.8 & 56.2 & 58.1 & 59.0 & 69.5 & 78.4 & 55.2 & 58.1 \\
Fang \etal~\cite{fang2017learning} & 50.1 & 54.3 & 57.0 & 57.1 & 66.6 & 73.3 & 53.4 & 55.7 \\
Cha \etal~\cite{cha2018pose} & 48.4 & \textbf{52.9} & 55.2 & 53.8 & 62.8 & 73.3 & 52.3 & 52.2 \\
Yang \etal~\cite{yang20183d} & 51.5 & 58.9 & \textbf{50.4} & 57.0 & 62.1 & \textbf{65.4} & \textbf{49.8} & 52.7 \\
\hline
FC baseline & 50.5 & 54.5 & 52.4 & 56.7 & 62.2 & 74.0 & 55.2 & 52.0 \\
RN-hier & 49.9 & 53.9 & 52.8 & 56.6 & 60.8 & 76.1 & 54.3 & 51.3 \\
RN & 49.7 & 54.0 & 52.0 & 56.4 & \textbf{60.9} & 74.1 & 53.4 & 51.1 \\
RN-FT & \textbf{49.4} & 54.3 & 51.6 & \textbf{55.0} & 61.0 & 73.3 & 53.7 & \textbf{50.0} \\
\hline\hline

Method & Sit & SitDown & Smoke & Wait & WalkD & Walk & WalkT & \textbf{Avg} \\
\hline
Pavlakos \etal~\cite{pavlakos2017coarse} & 83.7 & 96.5 & 71.7 & 65.8 & 74.9 & 59.1 & 63.2 & 71.9 \\
Tekin \etal~\cite{Tekin_2017_ICCV} & 70.1 & 107.3 & 69.3 & 70.3 & 74.3 & 51.8 & 63.2 & 69.7 \\
Zhou \etal~\cite{zhou2017towards} & 75.2 & 111.6 & 64.2 & 66.1 & \textbf{51.4} & 63.2 & 55.3 & 64.9 \\
Martinez \etal~\cite{Martinez_2017_ICCV} & 74.0 & 94.6 & 62.3 & 59.1 & 65.1 & 49.5 & 52.4 & 62.9 \\
Fang \etal~\cite{fang2017learning} & 72.8 & 88.6 & 60.3 & 57.7 & 62.7 & 47.5 & 50.6 & 60.4 \\
Cha \etal~\cite{cha2018pose} & 71.0 & 89.9 & 58.2 & \textbf{53.6} & 61.0 & 43.2 & 50.0 & 58.8 \\
Yang \etal~\cite{yang20183d} & 69.2 & \textbf{85.2} & \textbf{57.4} & 58.4 & 60.1 & \textbf{43.6} & \textbf{47.7} & \textbf{58.6} \\
\hline
FC baseline & 70.0 & 90.8 & 58.7 & 56.8 & 60.4 & 46.3 & 52.2 & 59.7 \\
RN-hier & 68.5 & 90.9 & 58.5 & 56.4 & 59.3 & 45.5 & 50.0 & 59.2 \\
RN & 69.3 & 90.4 & 58.1 & 56.4 & 59.5 & 45.6 & 50.6 & 59.0 \\
RN-FT & \textbf{68.5} & 88.7 & 58.6 & 56.8 & 57.8 & 46.2 & 48.6 & \textbf{58.6} \\
\hline
\end{tabular}
}
\end{center}
\caption{MPJPE (in mm) on Human 3.6M dataset under Protocol 1.}
\label{tab1}
\end{table}

The MPJPE \nj{of various algorithms using} Protocol 1 is provided in Table~\ref{tab1}. It can be seen that the baseline network already outperforms most of the existing methods, which validates the superiority of the proposed residual modules. The \nj{relational} networks are trained without applying relational dropouts. The proposed relational network (\textit{RN}) gains 0.7 mm improvements over the baseline \nj{on average}, and it is further improved when the network is finetuned on each sequence (\textit{RN-FT}), which \nj{achieves} state-of-the-art performance. Therefore, it is verified that capturing relations between different groups of joints improves the pose estimation performance despite its simpler structure and training procedures than the compared methods. Hierarchical relational networks (\textit{RN-hier}) does not outperform \textit{RN} although it has bigger number of parameters than \textit{RN}. We conjecture the reason \nj{to be} that it is hard to capture the useful relations in \nj{a} small number of joints which leads to \nj{output poorer} features than \nj{the ones} using the raw 2D positions.

\begin{table}[t]
\begin{center}
\adjustbox{width=0.8\linewidth}{
\begin{tabular}{|l|c @{\hskip 0.3em} c @{\hskip 0.3em} c @{\hskip 0.3em} c @{\hskip 0.3em} c @{\hskip 0.3em} c @{\hskip 0.3em} c @{\hskip 0.3em} c|}
\hline
Method & Direct & Discuss & Eat & Greet & Phone & Photo & Pose & Purchase \\
\hline
Moreno-Noguer~\cite{moreno20173d} & 66.1 & 61.7 & 84.5 & 73.7 & 65.2 & 67.2 & 60.9 & 67.3 \\
Martinez \etal~\cite{Martinez_2017_ICCV} & 39.5 & 43.2 & 46.4 & 47.0 & 51.0 & 56.0 & 41.4 & 40.6 \\
Fang \etal~\cite{fang2017learning} & 38.2 & 41.7 & 43.7 & 44.9 & 48.5 & 55.3 & 40.2 & 38.2 \\
Cha \etal~\cite{cha2018pose} & 39.6 & 41.7 & 45.2 & 45.0 & 46.3 & 55.8 & 39.1 & 38.9 \\
Yang \etal~\cite{yang20183d} & 26.9 & 30.9 & 36.3 & 39.9 & 43.9 & 47.4 & 28.8 & 29.4 \\
\hline
FC baseline & 43.3 & 45.7 & 44.2 & 48.0 & 51.0 & 56.8 & 44.3 & 41.1 \\
RN-hier & 42.5 & 44.9 & 44.2 & 47.4 & 49.1 & 57.4 & 43.9 & 40.5 \\
RN & 42.4 & 45.2 & 44.2 & 47.5 & 49.5 & 56.4 & 43.0 & 40.5 \\
RN-FT & 38.3 & 42.5 & 41.5 & 43.3 & 47.5 & 53.0 & 39.3 & 37.1 \\
\hline\hline

Method & Sit & SitDown & Smoke & Wait & WalkD & Walk & WalkT & \textbf{Avg} \\
\hline
Moreno-Noguer~\cite{moreno20173d} & 103.5 & 74.6 & 92.6 & 69.6 & 71.5 & 78.0 & 73.2 & 74.0 \\
Martinez \etal~\cite{Martinez_2017_ICCV} & 56.5 & 69.4 & 49.2 & 45.0 & 49.5 & 38.0 & 43.1 & 47.7 \\
Fang \etal~\cite{fang2017learning} & 54.5 & 64.4 & 47.2 & 44.3 & 47.3 & 36.7 & 41.7 & 45.7 \\
Cha \etal~\cite{cha2018pose} & 55.0 & 67.2 & 45.9 & 42.0 & 47.0 & 33.1 & 40.5 & 45.7\\
Yang \etal~\cite{yang20183d} & 36.9 & 58.4 & 41.5 & 30.5 & 29.5 & 42.5 & 32.2 & 37.7 \\
\hline
FC baseline & 57.0 & 68.8 & 49.2 & 45.3 & 50.5 & 38.2 & 45.0 & 48.9 \\
RN-hier & 56.7 & 68.5 & 48.5 & 44.7 & 49.4 & 37.0 & 43.1 & 48.1 \\
RN & 56.8 & 68.4 & 48.4 & 44.7 & 49.8 & 37.6 & 44.1 & 48.2 \\
RN-FT & 54.1 & 64.3 & 46.0 & 42.0 & 44.8 & 34.7 & 38.7 & 45.0 \\
\hline
\end{tabular}
}
\end{center}
\caption{MPJPE on Human 3.6M dataset under Protocol 2.}
\label{tab2}
\end{table}

The MPJPE \nj{using the alignment} Protocol 2 is provided in Table~\ref{tab2}. When shape aligning via Procrustes analysis is applied, our method \nj{RN-FT} showed superior performance to the existing methods except~\cite{yang20183d}.

\begin{table}[t]
\begin{center}
\adjustbox{width=0.8\linewidth}{
\begin{tabular}{|l|c c c c| c c c c|}
\hline
\multirow{ 2}{*}{Method} & \multicolumn{4}{|c|}{Protocol 1} & \multicolumn{4}{|c|}{Protocol 2} \\
 & None & Rand 2 & L Arm & R Leg & None & Rand 2 & L Arm & R Leg \\
 \hline
Moreno-Noguer~\cite{moreno20173d} & - & - & - & - & 74.0 & 106.8 & 109.4 & 100.2 \\
FC baseline & 59.7 & 256.1 & 213.9 & 222.7 & 48.9 & 192.3 & 153.8 & 155.7 \\
FC-drop & 68.6 & 241.6 & 98.1 & 90.6 & 52.3 & 159.7 & 82.0 & 70.2 \\
RN & \textbf{59.0} & 540.2 & 314.1 & 332.8 & 48.2 & 280.7 & 225.8 & 214.1 \\
RN-drop & 59.3 & 218.7 & \textbf{73.8} & 70.6 & \textbf{45.5} & 145.3 & \textbf{62.7} & \textbf{55.0} \\
RN-hier-drop & 59.7 & \textbf{65.9} & 74.5 & \textbf{70.4} & 45.6 & \textbf{51.4} & 63.0 & 55.2 \\
 \hline
\end{tabular}
}
\end{center}
\caption{MPJPE on Human 3.6M dataset with various types of missing joints.}
\label{tab3}
\end{table}

Next, we discuss the effectiveness of the relational dropout for the case of missing joints. MPJPE for all sequences with various types of missing joints are measured and provided in Table~\ref{tab3}. We simulated 3 types of missing joints following~\cite{moreno20173d}, \nj{which are} 2 random joints (Rand 2), left arm (L Arm), and right leg (R Leg). We consider 3 \nj{missing joints} for the latter 2 cases including shoulder or hip joints. Note that \cite{moreno20173d} used different training schemes for experiments on missing joints where six subjects \nj{were} used for training. For the baseline method that can be applied to the fully connected network, we assign zero to the value of input 2D joints with the probability of 0.1, which is denoted as \textit{FC-drop}. It imposes robustness to the missing joints compared to the \textit{FC baseline} in which random drop is not applied. When relational dropout is applied to the relational network (\textit{RN-drop}), the model outperforms \textit{FC-drop} in all cases. The model successfully estimates 3D pose when one of the groups in the relational network is missing. Therefore, it shows smaller MPJPE when \nj{the} left arm or \nj{the} right leg is not visible. However, when two joints \nj{belonging} to different groups \nj{are missing}, \nj{the} two groups are dropped at the same time, which is not simulated during the training. Thus, \textit{RN-drop} shows poor performance for the case that random two joints are missing. This problem can be handled when relational dropout is applied to the hierarchical relational network. When one joint is missing in a group, relational dropout is applied to hierarchical relational unit within the group. In the case that two or more joints are missing in a group, relational dropout is applied to the group. This model (\textit{RN-hier-drop}) showed impressive performance in all types of missing \nj{joints}. Another advantage of the relational dropout is \nj{that} it does not degrade the performance of the case \nj{of all-visible joints}. It can be inferred that the robustness on missing joints increases as various combinations of missing joints are simulated during the training.

\begin{figure}[t]
\centering
 \begin{minipage}{0.15\textwidth}
 \centering
     2D inputs
 \end{minipage}
 \begin{minipage}{0.15\textwidth}
 \centering
 	RN
 \end{minipage}
 \begin{minipage}{0.15\textwidth}
 \centering
     FC-drop
 \end{minipage}
 \begin{minipage}{0.15\textwidth}
 \centering
     RN-drop
 \end{minipage}
 \begin{minipage}{0.15\textwidth}
 \centering
 	RN-hier-drop
 \end{minipage}
 \begin{minipage}{0.15\textwidth}
 \centering
     GT
 \end{minipage}

 \begin{minipage}{0.15\textwidth}
 \centering
     \includegraphics[width=\linewidth]{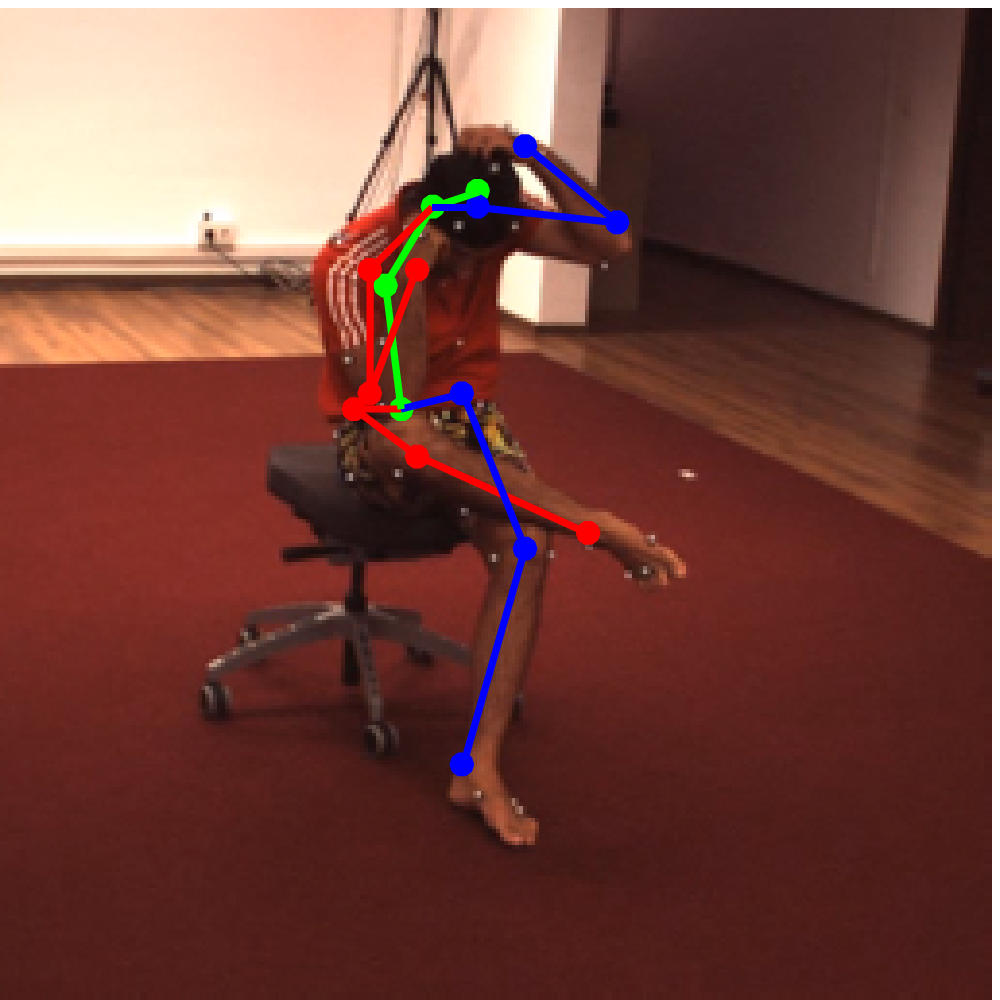}
 \end{minipage}
 \begin{minipage}{0.15\textwidth}
 \centering
     \includegraphics[width=\linewidth]{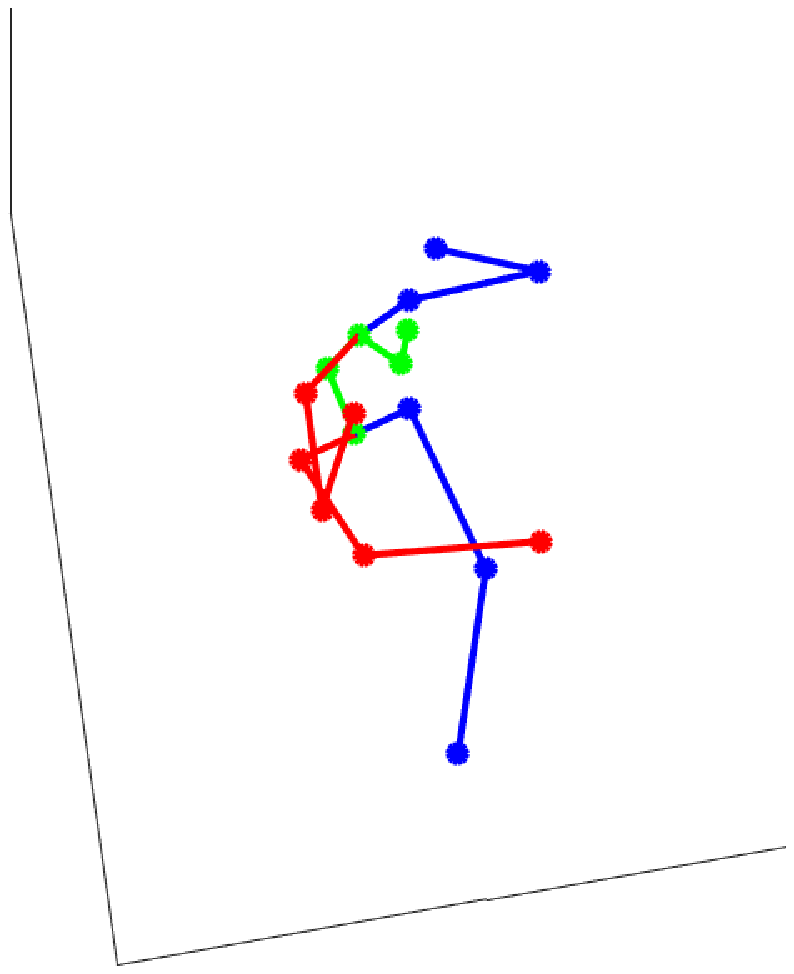}
 \end{minipage}
 \begin{minipage}{0.15\textwidth}
 \centering
     \includegraphics[width=\linewidth]{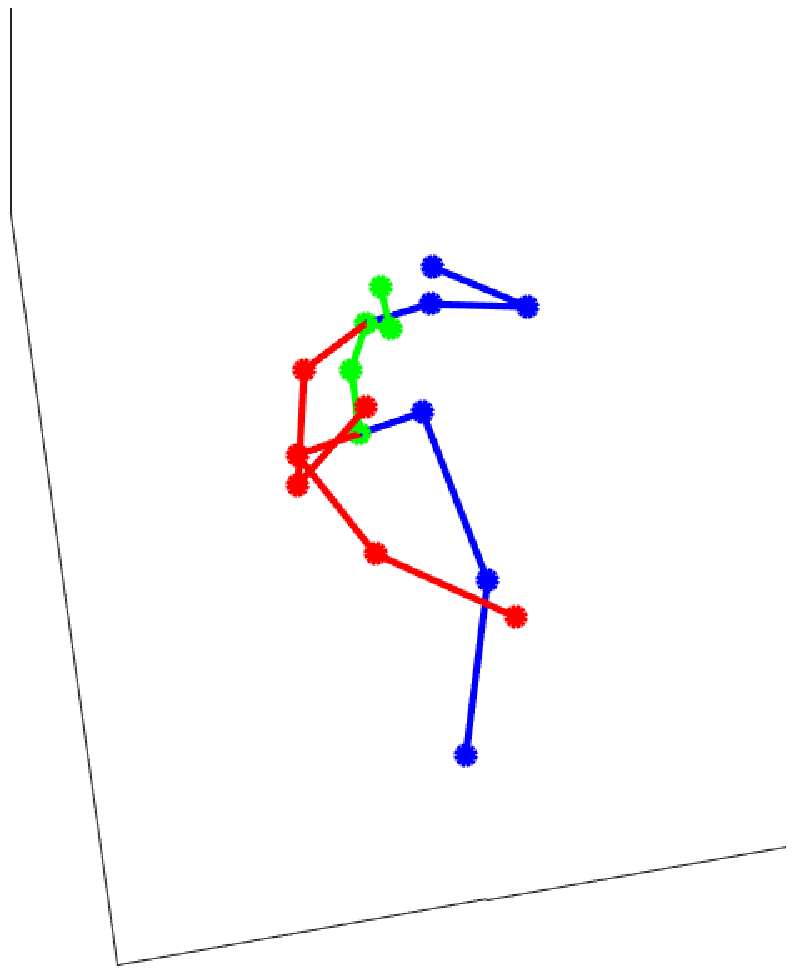}
 \end{minipage}
 \begin{minipage}{0.15\textwidth}
 \centering
     \includegraphics[width=\linewidth]{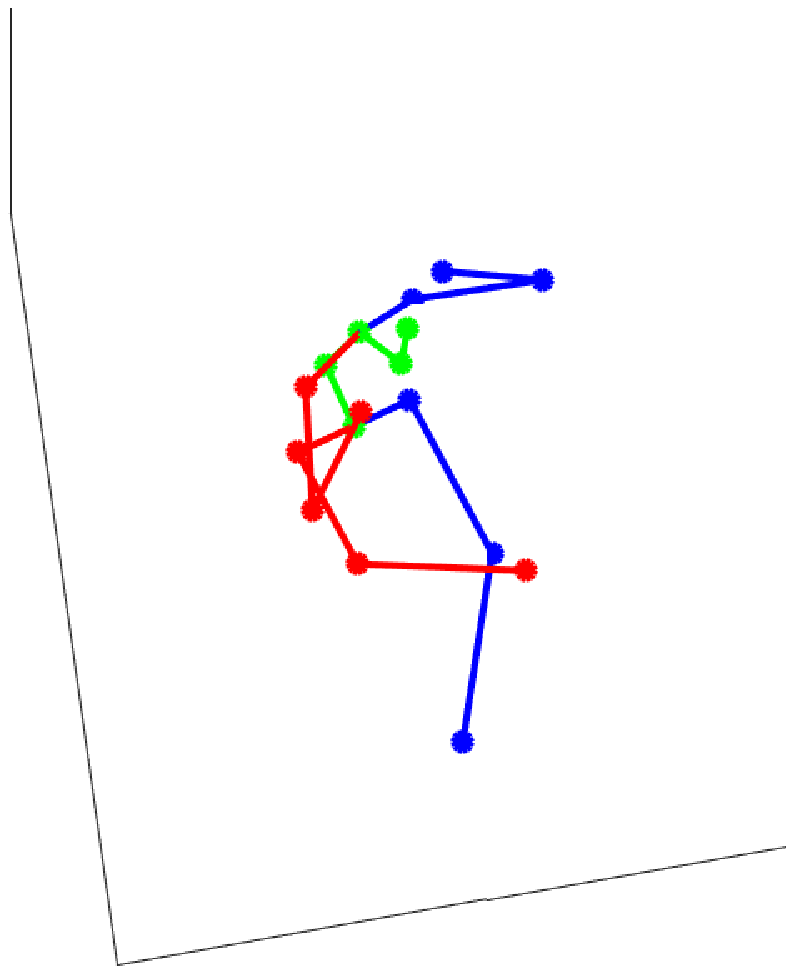}
 \end{minipage}
 \begin{minipage}{0.15\textwidth}
 \centering
     \includegraphics[width=\linewidth]{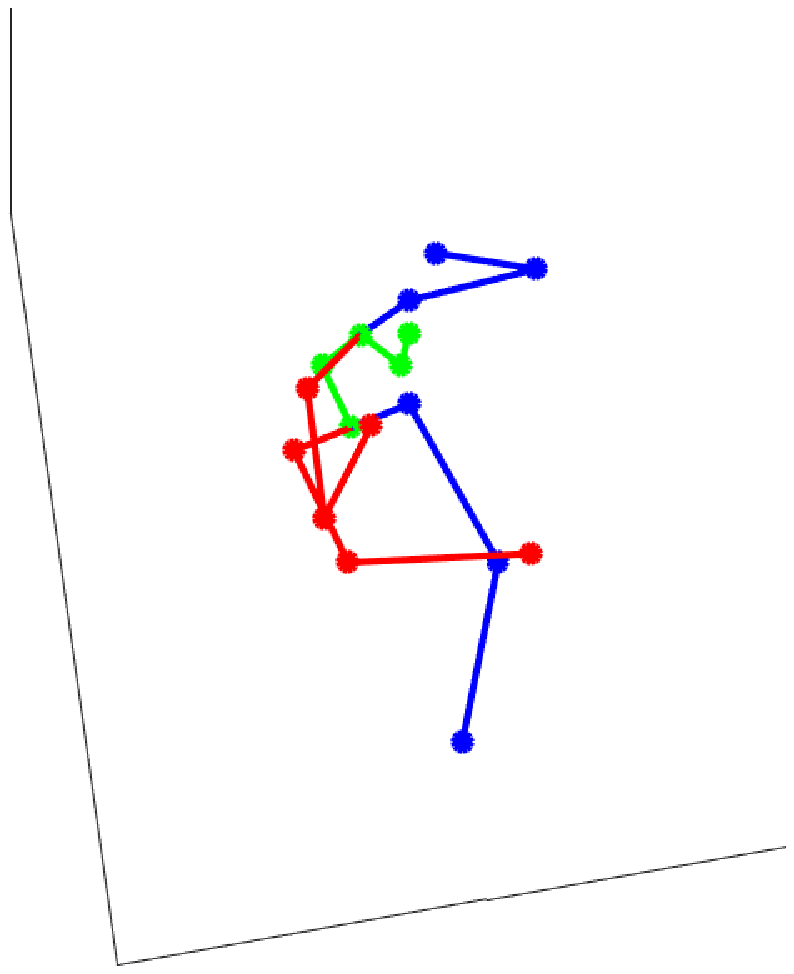}
 \end{minipage}
 \begin{minipage}{0.15\textwidth}
 \centering
     \includegraphics[width=\linewidth]{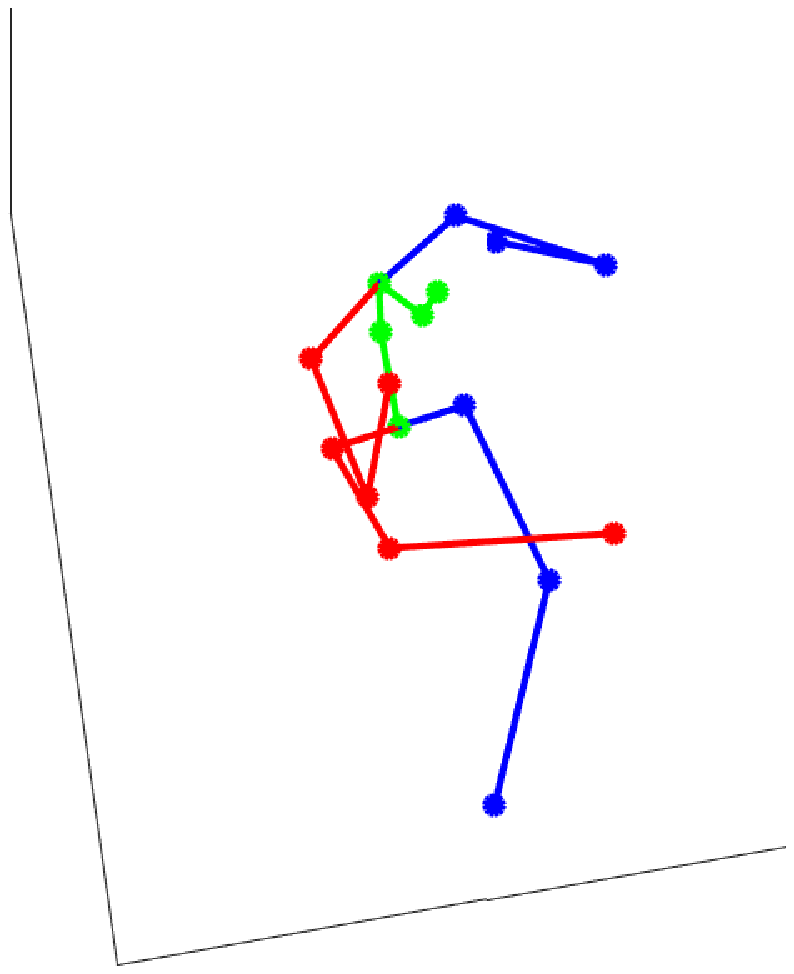}
 \end{minipage}

 \begin{minipage}{0.15\textwidth}
 \centering
     \includegraphics[width=\linewidth]{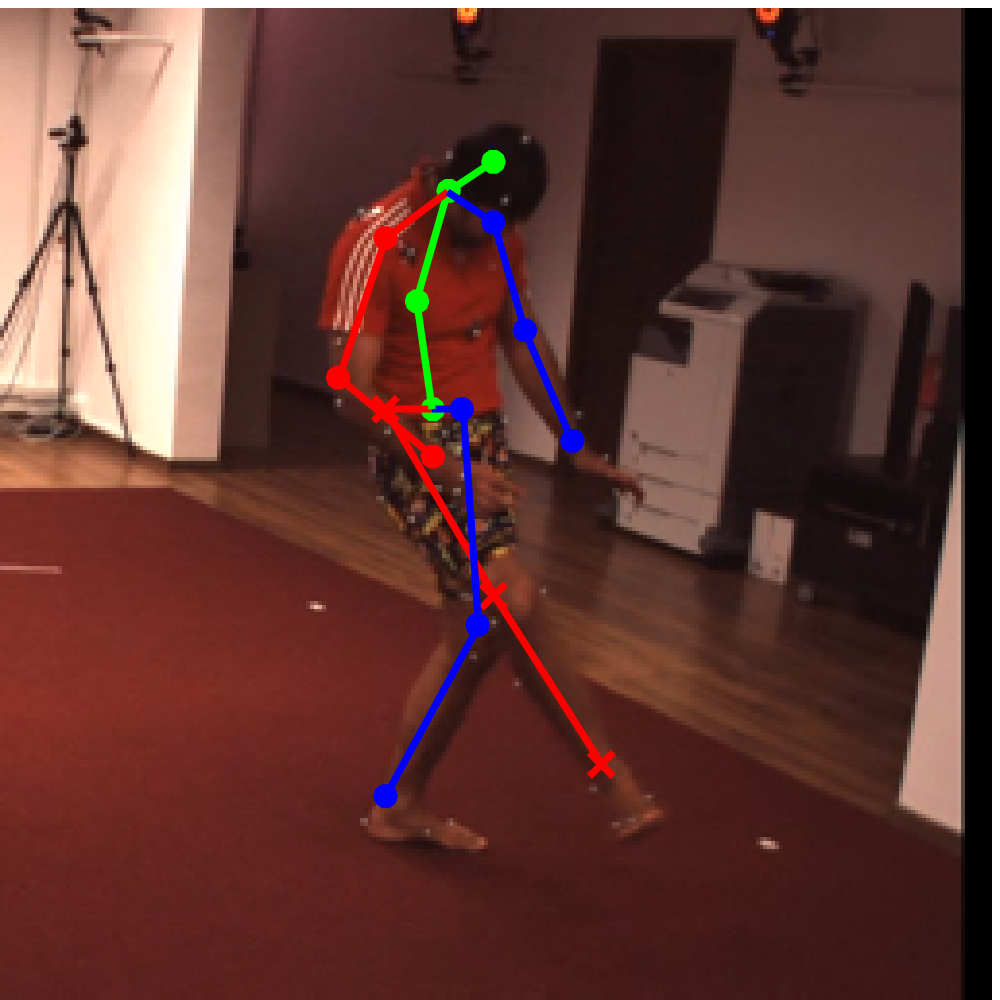}
 \end{minipage}
 \begin{minipage}{0.15\textwidth}
 \centering
     \includegraphics[width=\linewidth]{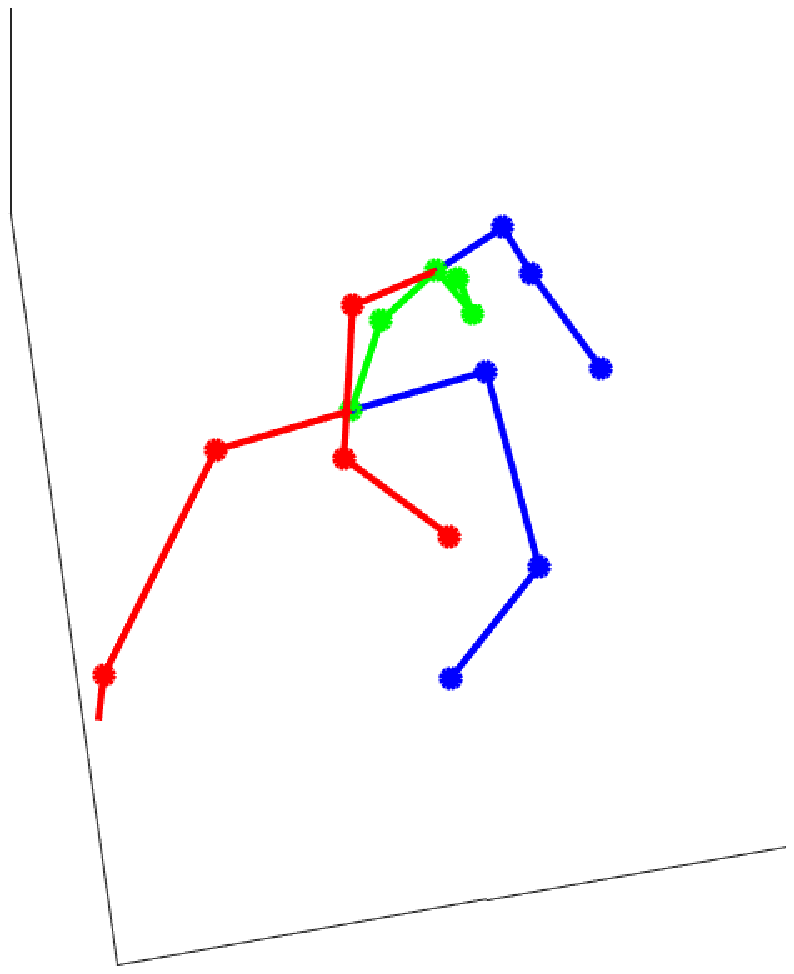}
 \end{minipage}
 \begin{minipage}{0.15\textwidth}
 \centering
     \includegraphics[width=\linewidth]{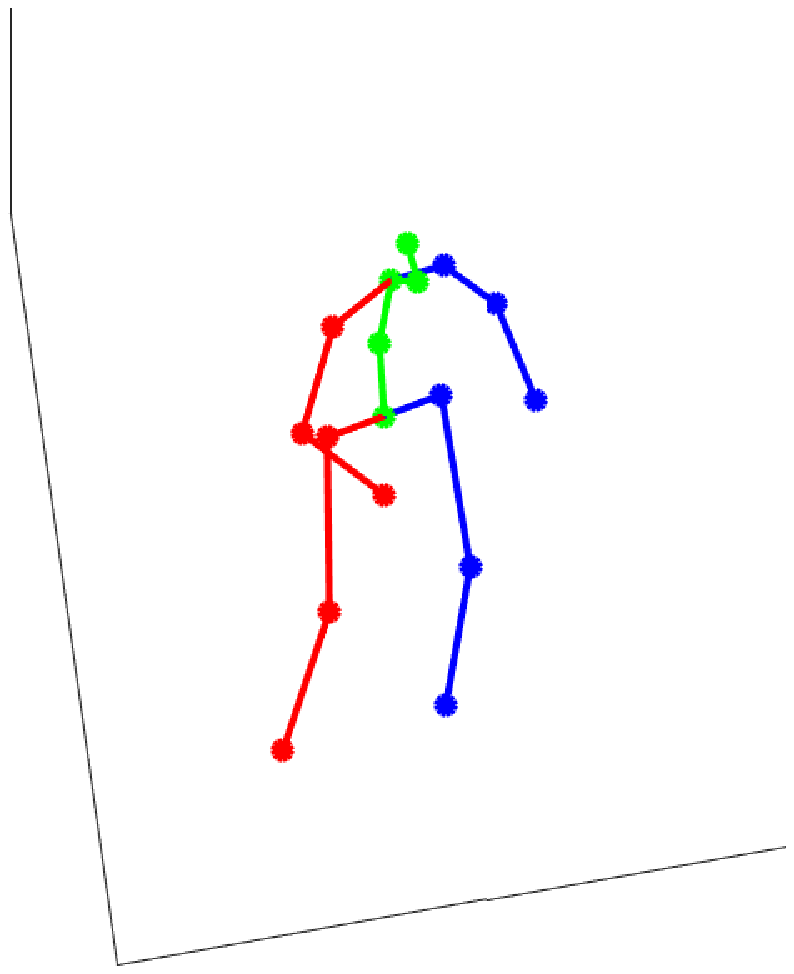}
 \end{minipage}
 \begin{minipage}{0.15\textwidth}
 \centering
     \includegraphics[width=\linewidth]{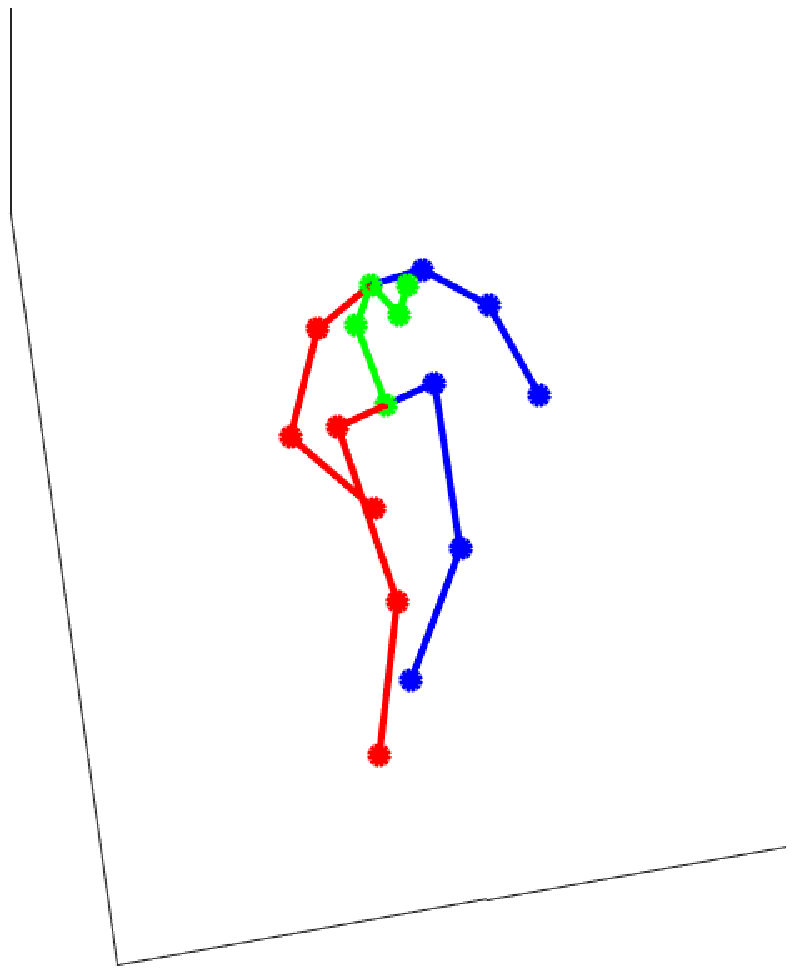}
 \end{minipage}
 \begin{minipage}{0.15\textwidth}
 \centering
     \includegraphics[width=\linewidth]{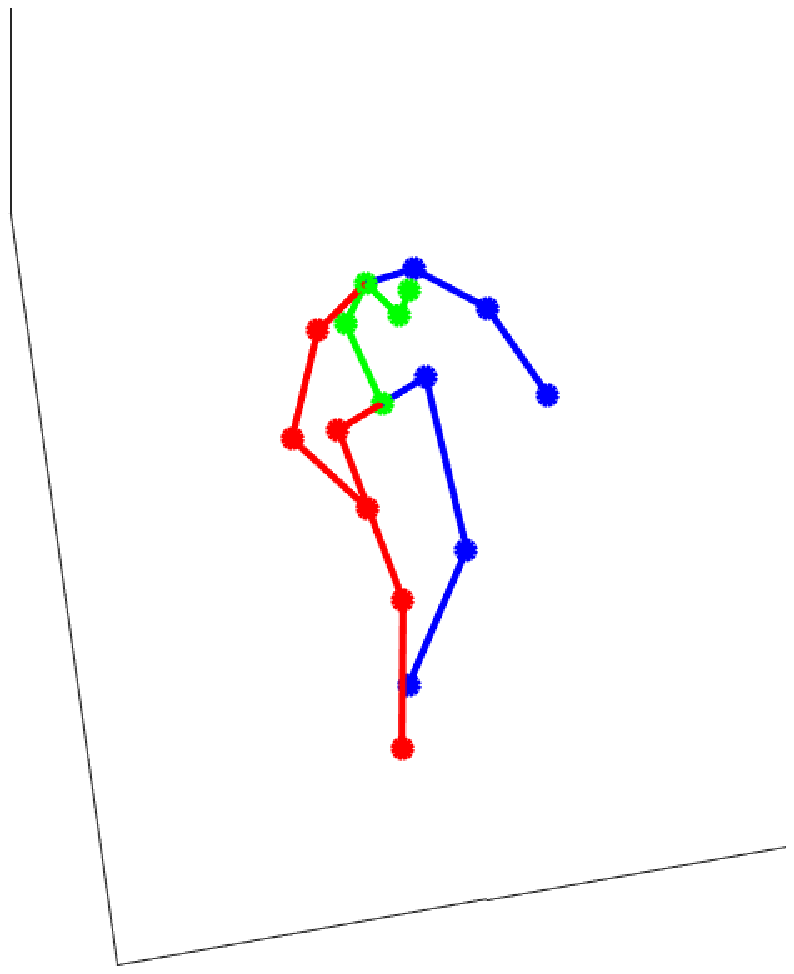}
 \end{minipage}
 \begin{minipage}{0.15\textwidth}
 \centering
     \includegraphics[width=\linewidth]{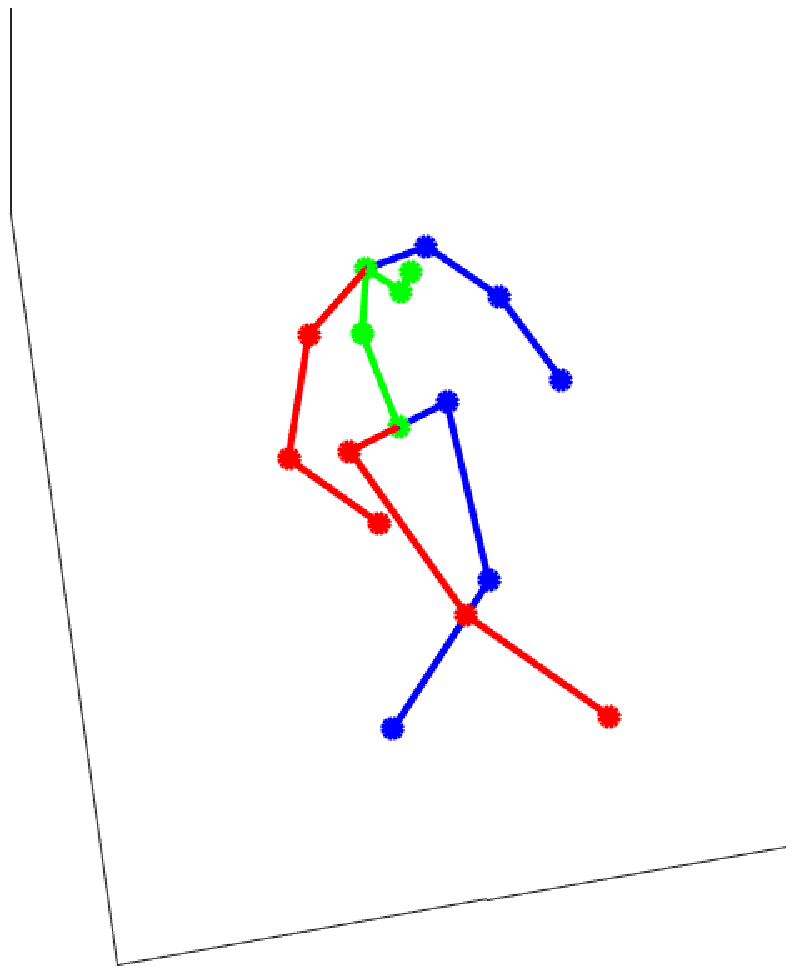}
 \end{minipage}

 \begin{minipage}{0.15\textwidth}
 \centering
     \includegraphics[width=\linewidth]{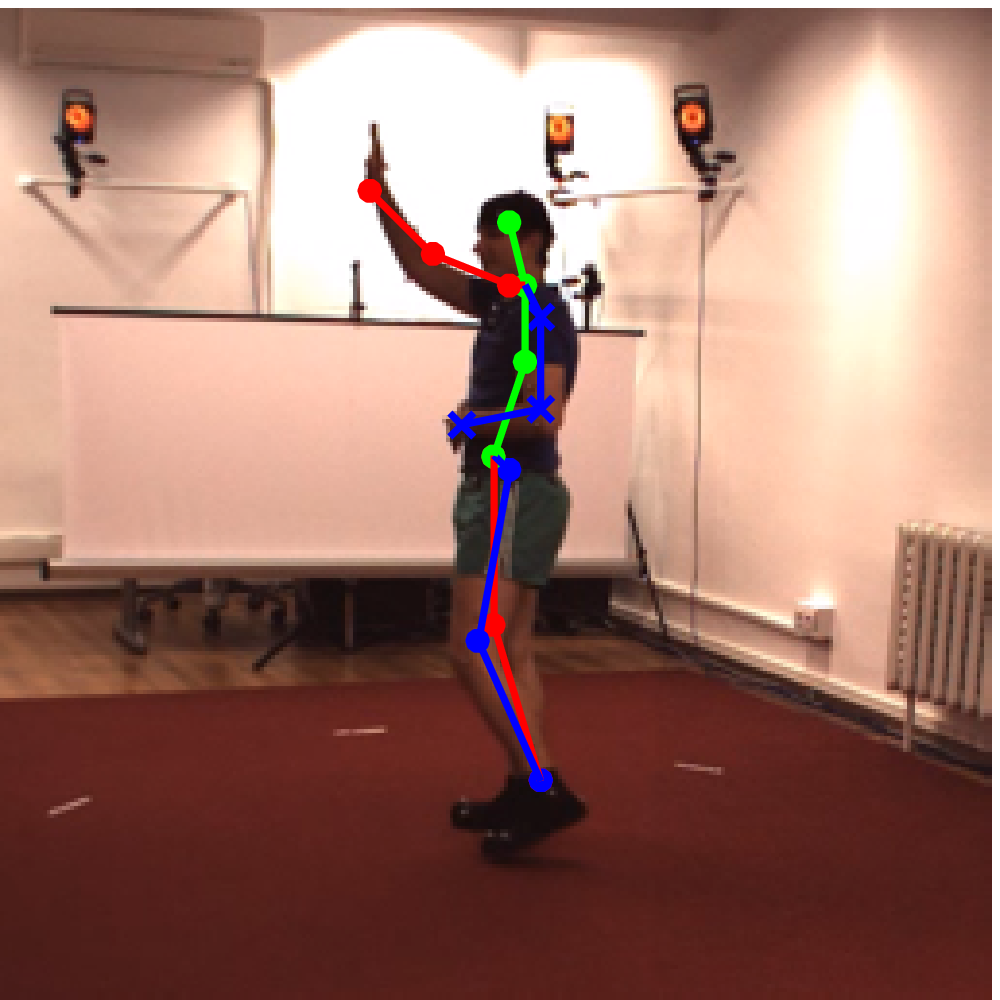}
 \end{minipage}
 \begin{minipage}{0.15\textwidth}
 \centering
     \includegraphics[width=\linewidth]{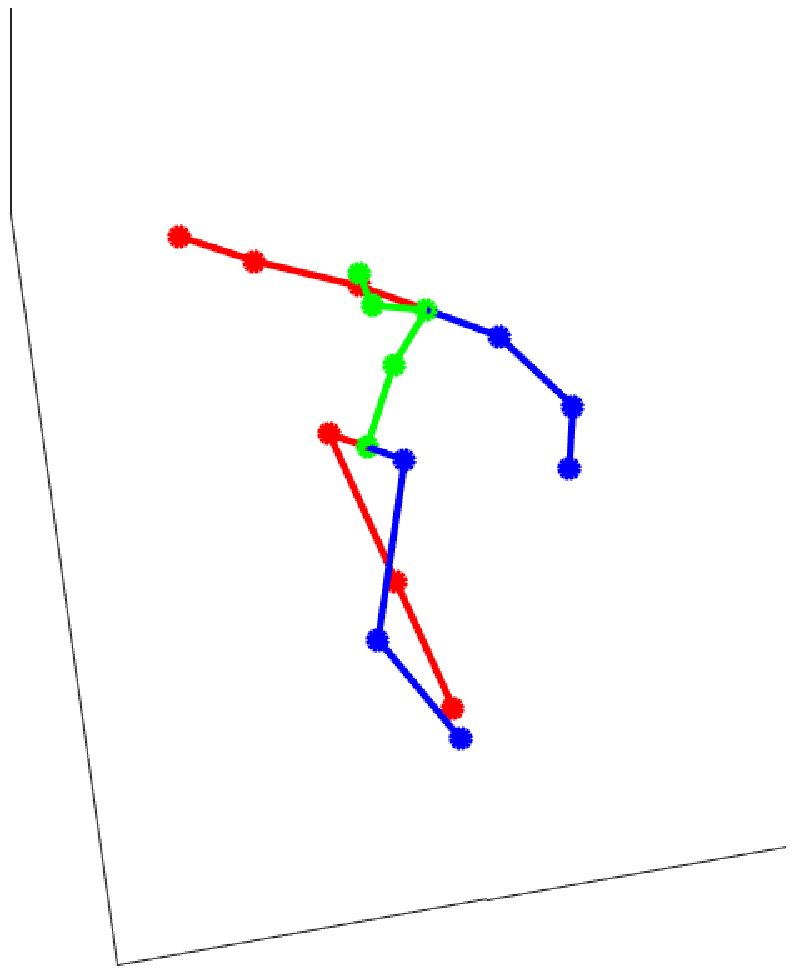}
 \end{minipage}
 \begin{minipage}{0.15\textwidth}
 \centering
     \includegraphics[width=\linewidth]{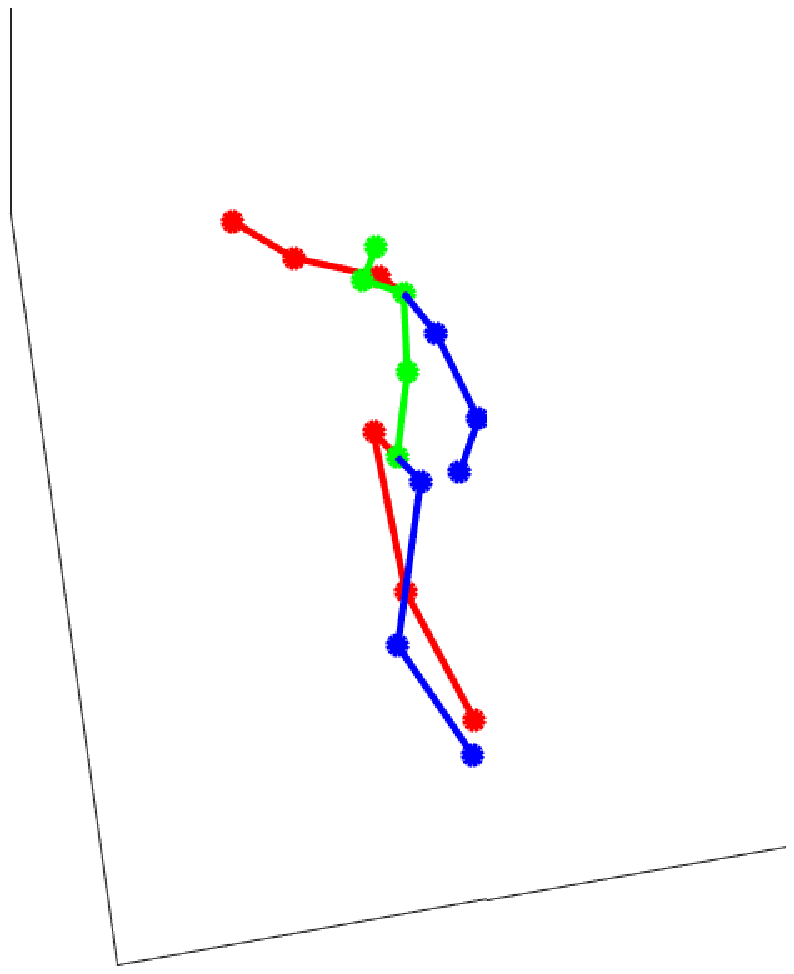}
 \end{minipage}
 \begin{minipage}{0.15\textwidth}
 \centering
     \includegraphics[width=\linewidth]{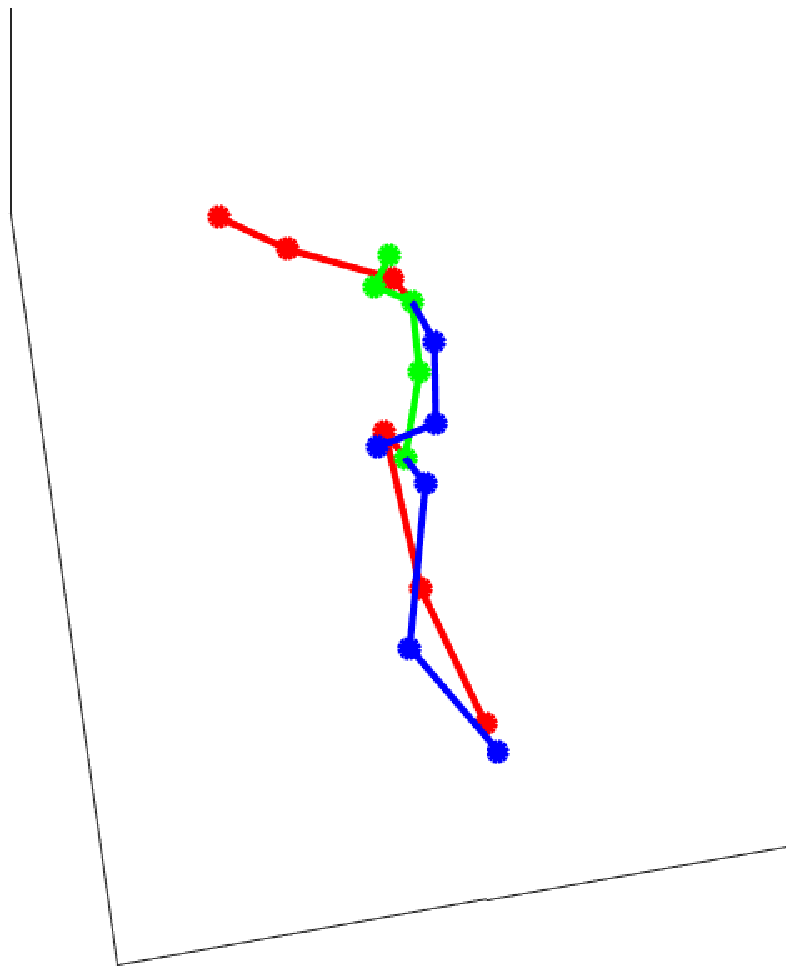}
 \end{minipage}
 \begin{minipage}{0.15\textwidth}
 \centering
     \includegraphics[width=\linewidth]{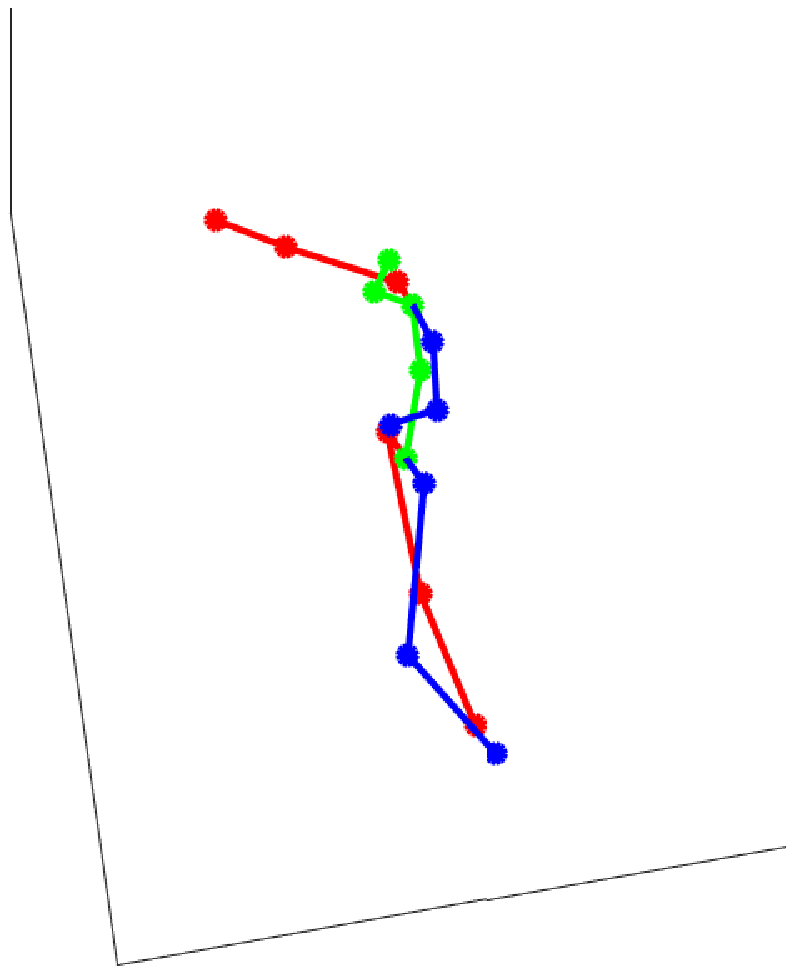}
 \end{minipage}
 \begin{minipage}{0.15\textwidth}
 \centering
     \includegraphics[width=\linewidth]{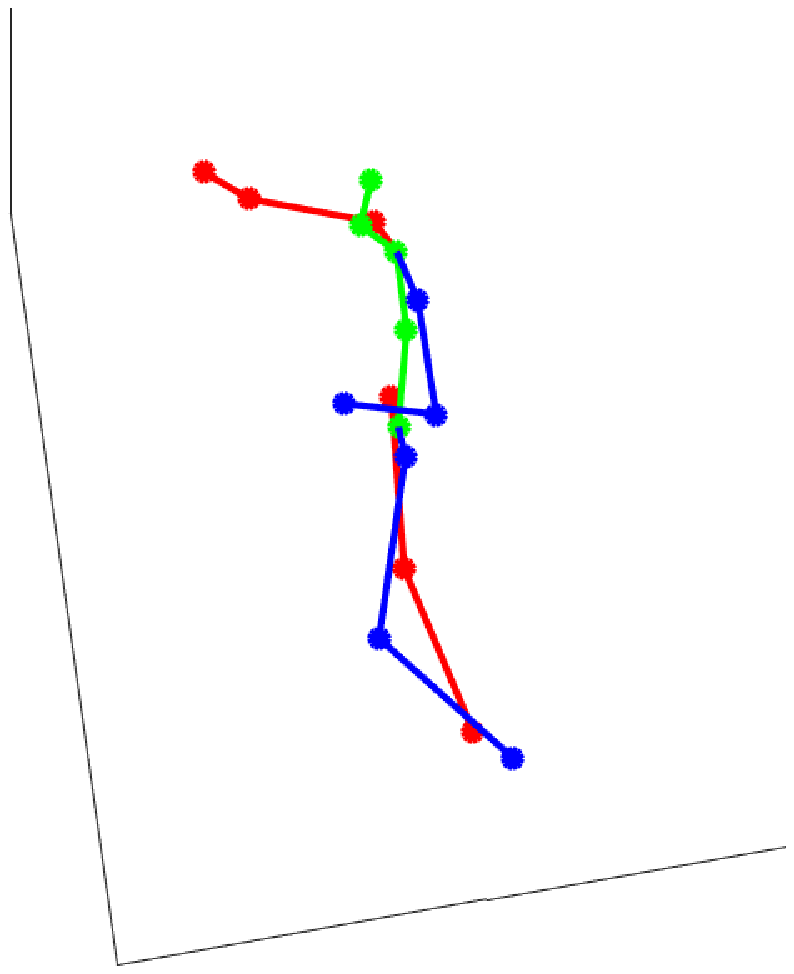}
 \end{minipage}

 \begin{minipage}{0.15\textwidth}
 \centering
     \includegraphics[width=\linewidth]{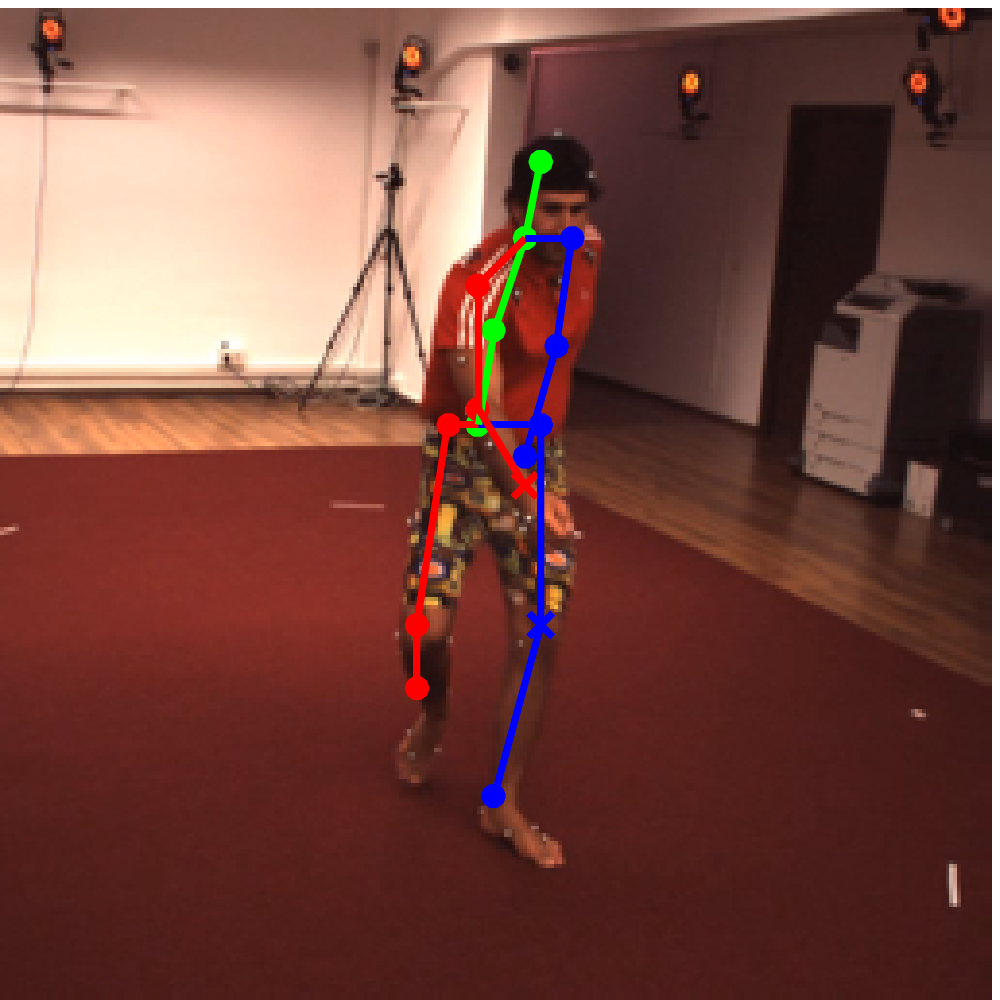}
 \end{minipage}
 \begin{minipage}{0.15\textwidth}
 \centering
     \includegraphics[width=\linewidth]{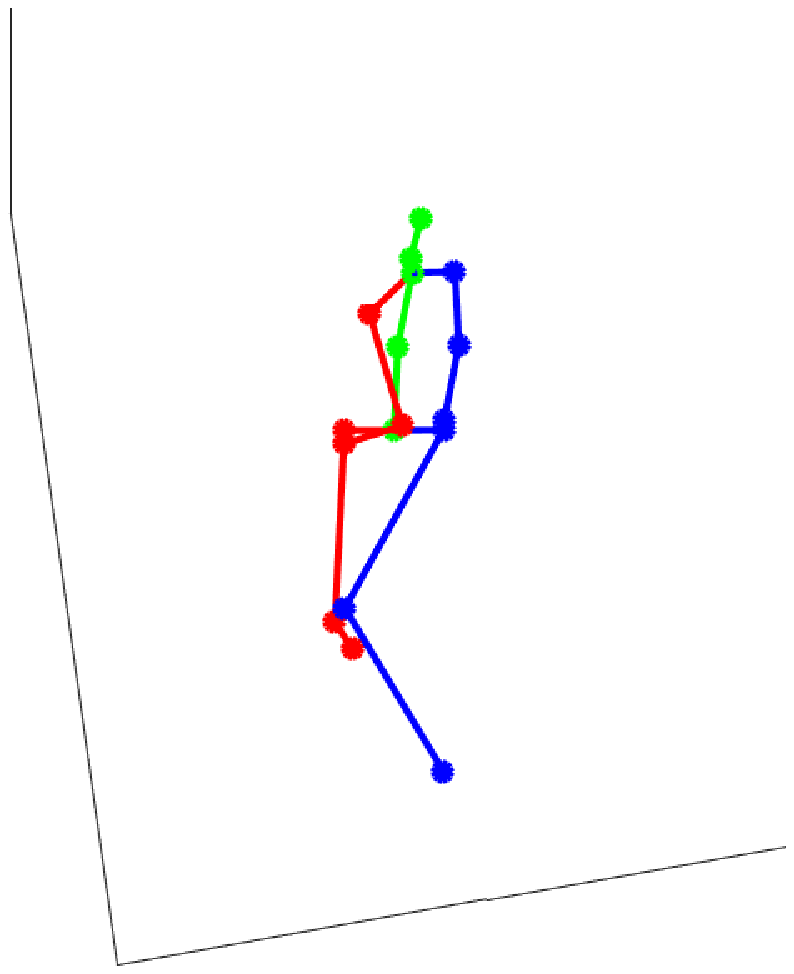}
 \end{minipage}
 \begin{minipage}{0.15\textwidth}
 \centering
     \includegraphics[width=\linewidth]{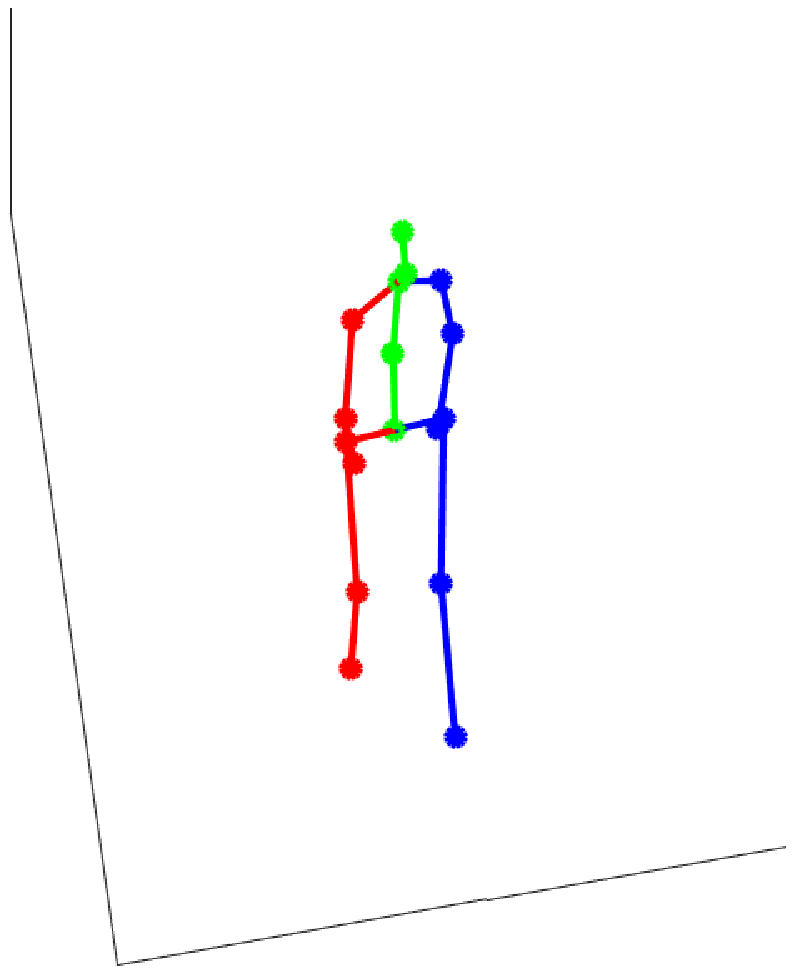}
 \end{minipage}
 \begin{minipage}{0.15\textwidth}
 \centering
     \includegraphics[width=\linewidth]{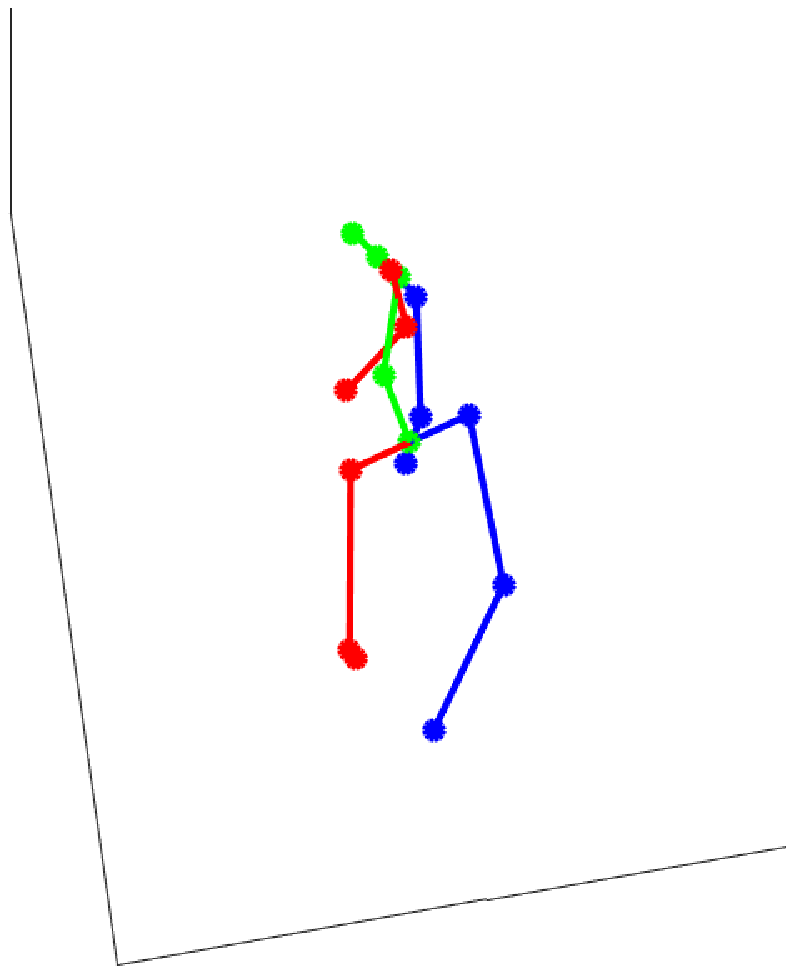}
 \end{minipage}
 \begin{minipage}{0.15\textwidth}
 \centering
     \includegraphics[width=\linewidth]{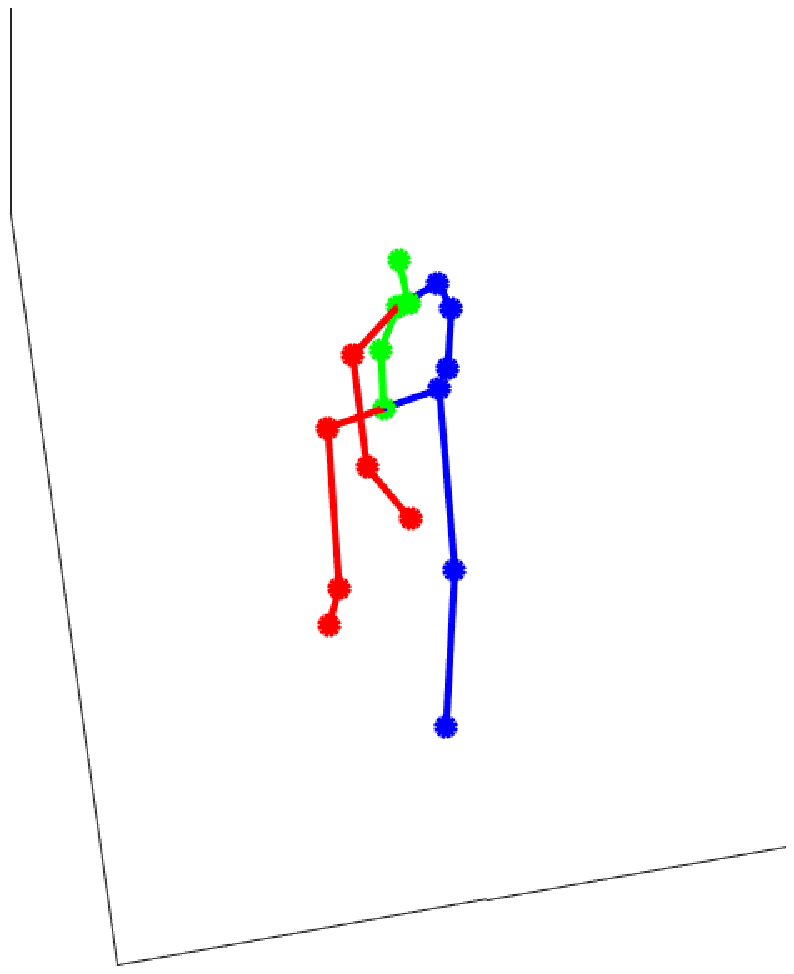}
 \end{minipage}
 \begin{minipage}{0.15\textwidth}
 \centering
     \includegraphics[width=\linewidth]{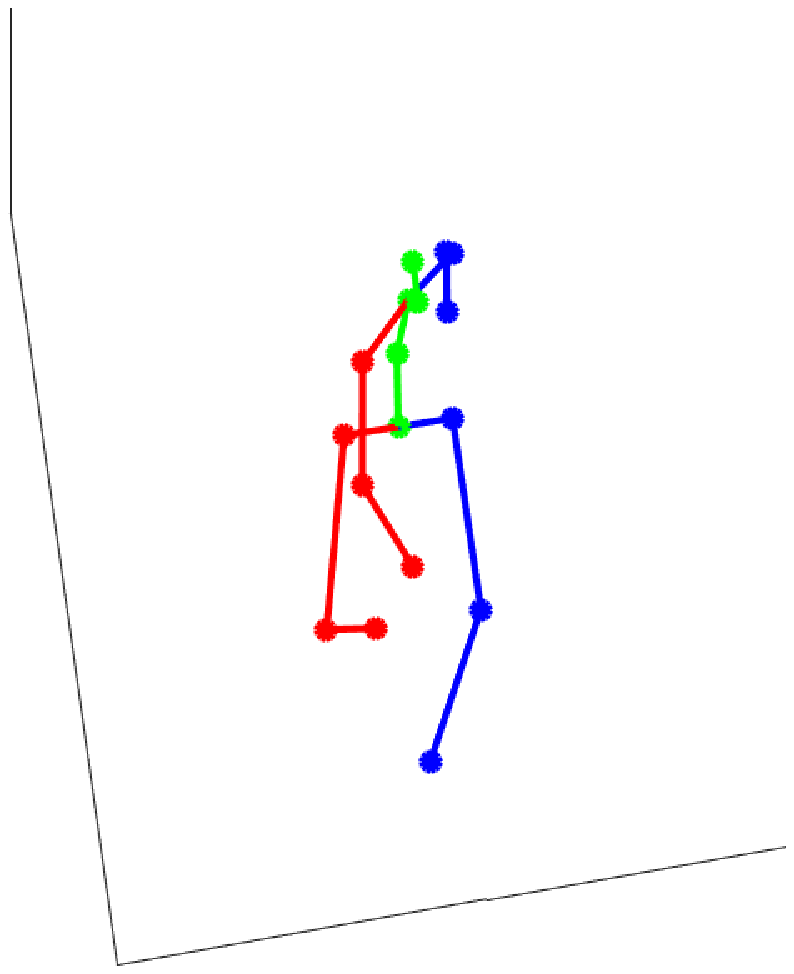}
 \end{minipage}

\caption{Qualitative results on Human 3.6M dataset in various cases of missing joints. For the 2D pose detection results, visible joints are marked as $\bullet$, and missing joints are marked as $\times$. Five groups are denoted as green (torso), red (right arm/leg) and blue (left arm/leg).}
\label{fig2}
\end{figure}

Qualitative results on Human 3.6M dataset are provided in Figure~\ref{fig2}. Each row simulates different cases of missing joints, none, right leg, left arm, and random 2 joints. The results of \textit{RN}, \textit{FC-drop}, \textit{RN-drop}, \textit{RN-hier-drop} is displayed with ground truth poses. When all joints are visible, all models generate similar poses that are close to the ground truth. On the other hand, \textit{RN} generates inaccurate poses when 2D inputs contain missing points. \textit{RN-drop} provides more accurate results than \textit{FC-drop}, but the model fails when joints of two different groups are missing. It can be seen that \textit{RN-hier-drop} outputs 3D poses that are similar to the ground truth poses in all cases. \sh{More results can be found on the supplementary materials.}

\begin{figure}[t]
\centering
 \begin{minipage}{0.15\textwidth}
 \centering
     2D inputs
 \end{minipage}
 \begin{minipage}{0.15\textwidth}
 \centering
 	RN
 \end{minipage}
 \begin{minipage}{0.15\textwidth}
 \centering
     RN-hier-drop
 \end{minipage}
 \begin{minipage}{0.15\textwidth}
 \centering
     2D inputs
 \end{minipage}
 \begin{minipage}{0.15\textwidth}
 \centering
 	RN
 \end{minipage}
 \begin{minipage}{0.15\textwidth}
 \centering
     RN-hier-drop
 \end{minipage}

 \begin{minipage}{0.15\textwidth}
 \centering
     \includegraphics[width=\linewidth]{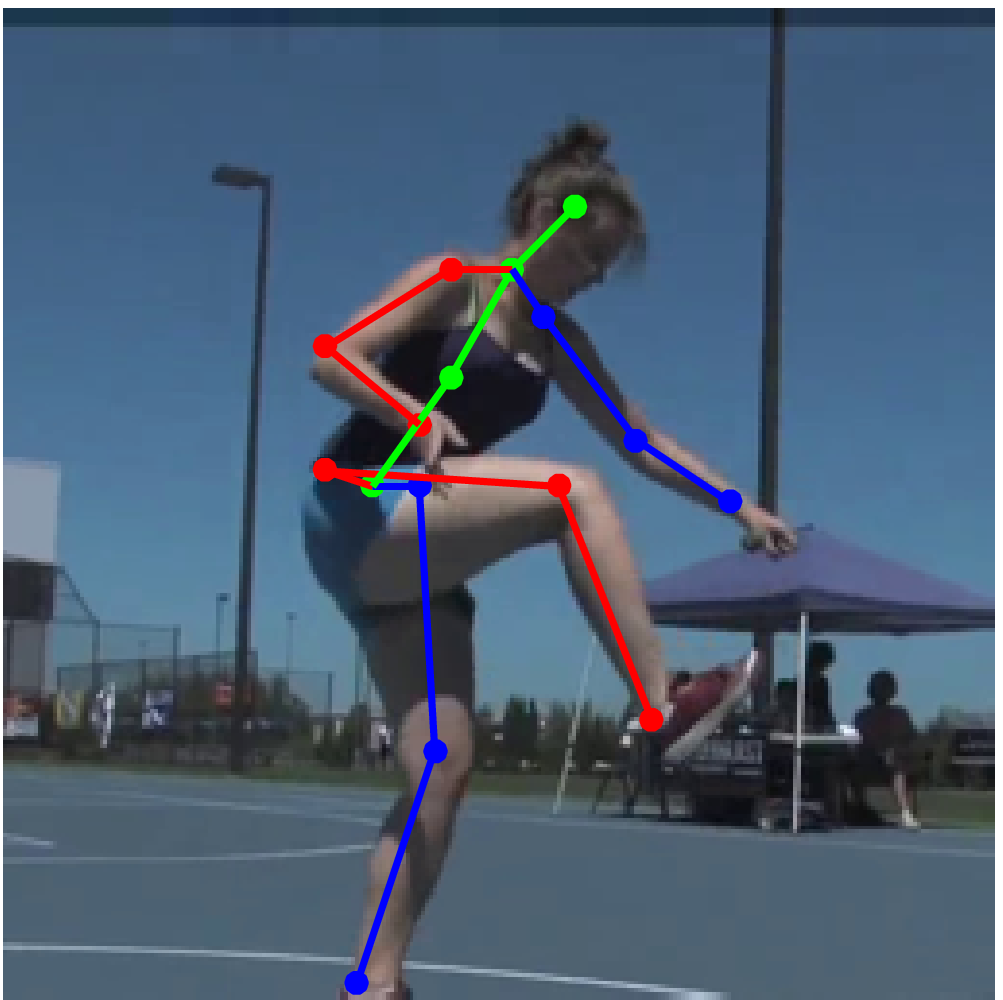}
 \end{minipage}
 \begin{minipage}{0.15\textwidth}
 \centering
     \includegraphics[width=\linewidth]{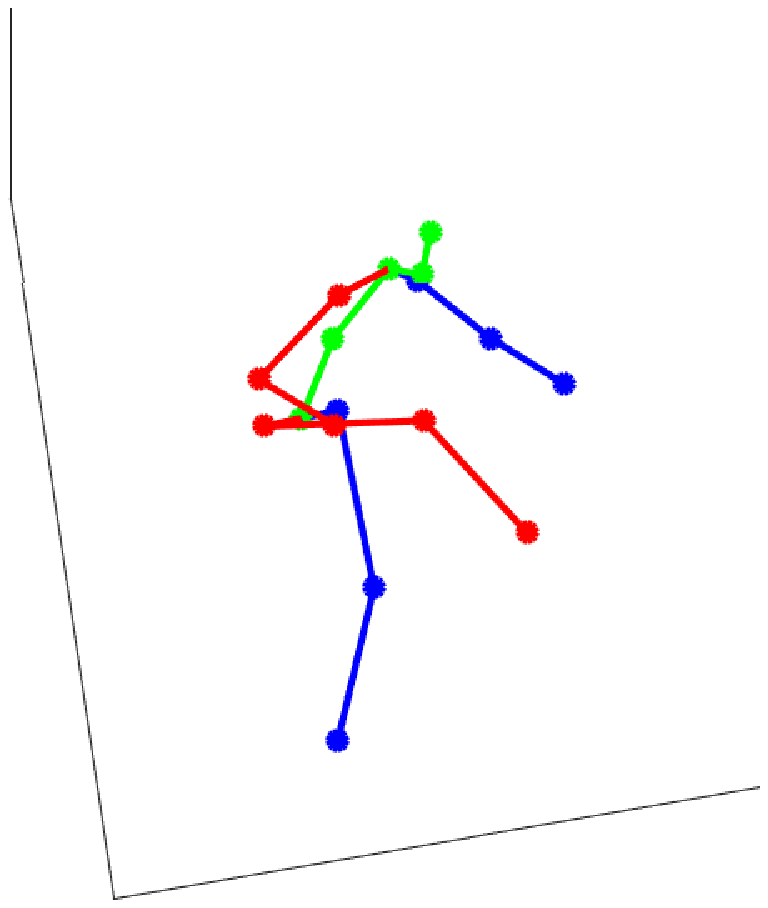}
 \end{minipage}
 \begin{minipage}{0.15\textwidth}
 \centering
     \includegraphics[width=\linewidth]{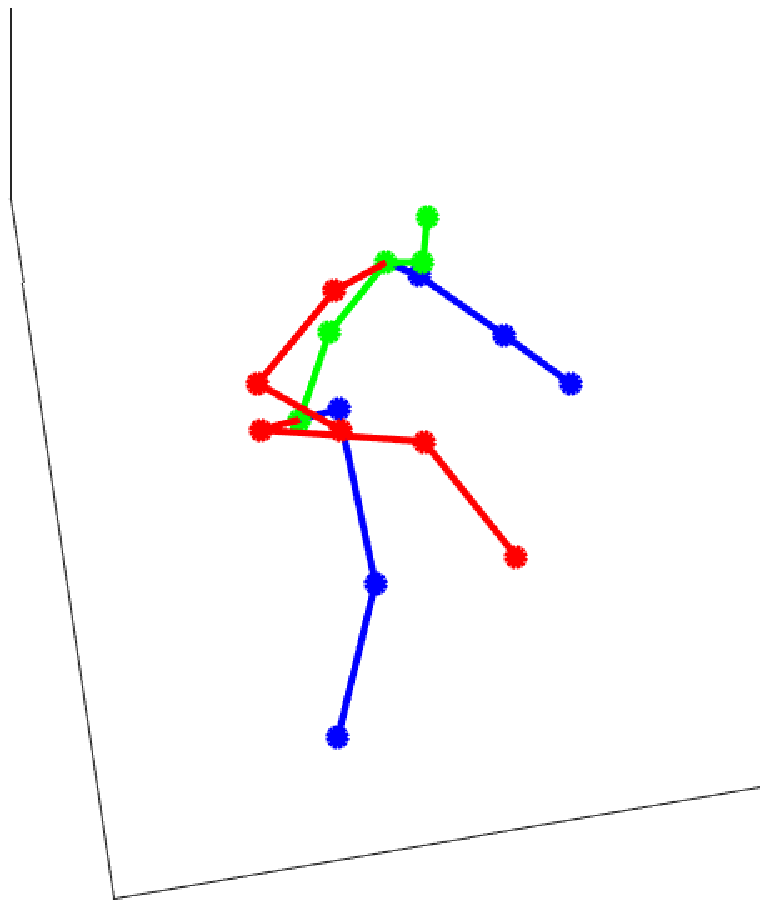}
 \end{minipage}
 \begin{minipage}{0.15\textwidth}
 \centering
     \includegraphics[width=\linewidth]{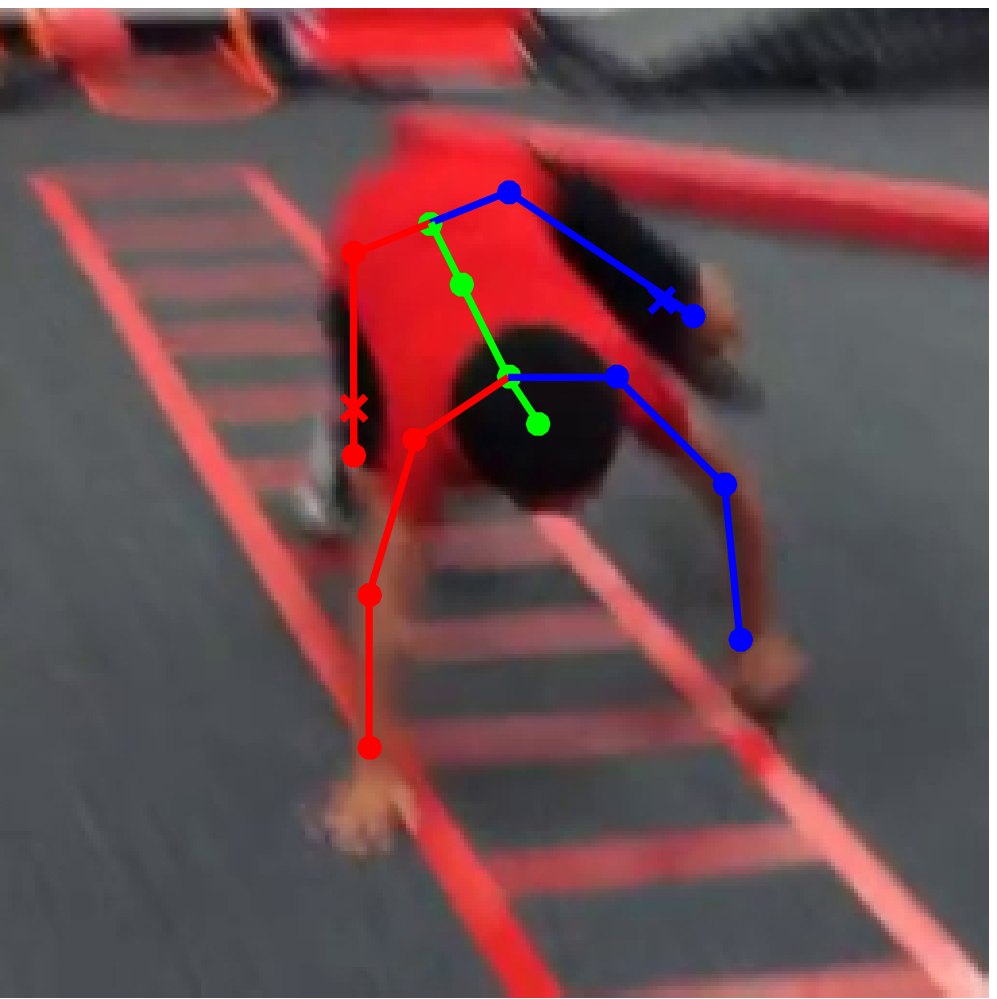}
 \end{minipage}
 \begin{minipage}{0.15\textwidth}
 \centering
     \includegraphics[width=\linewidth]{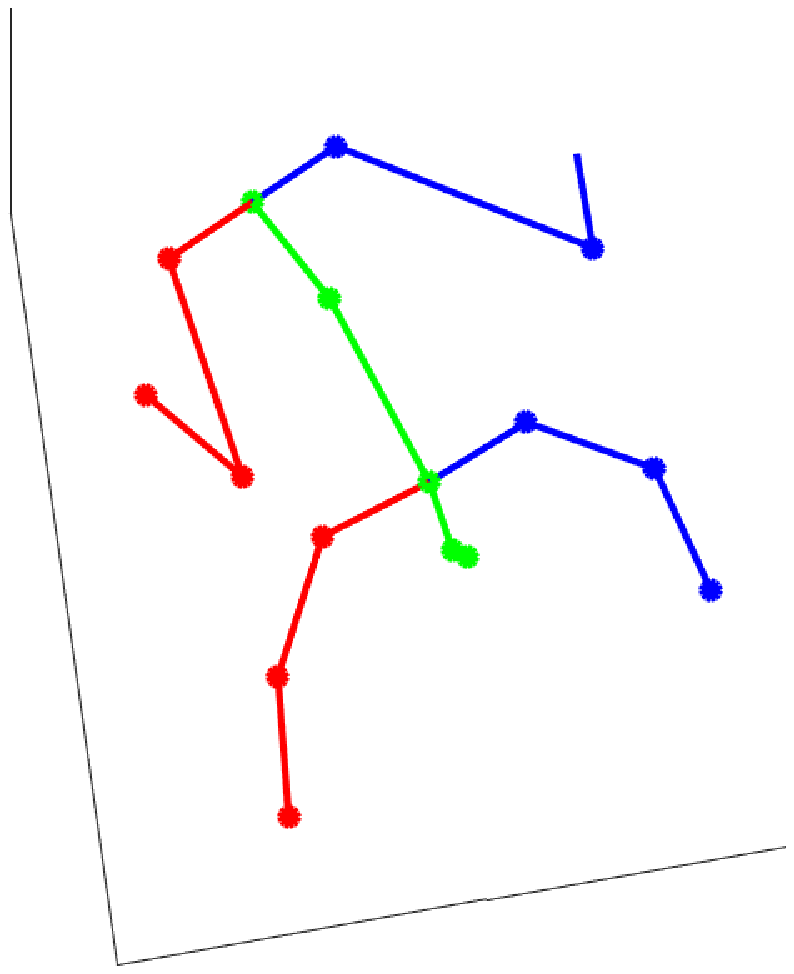}
 \end{minipage}
 \begin{minipage}{0.15\textwidth}
 \centering
     \includegraphics[width=\linewidth]{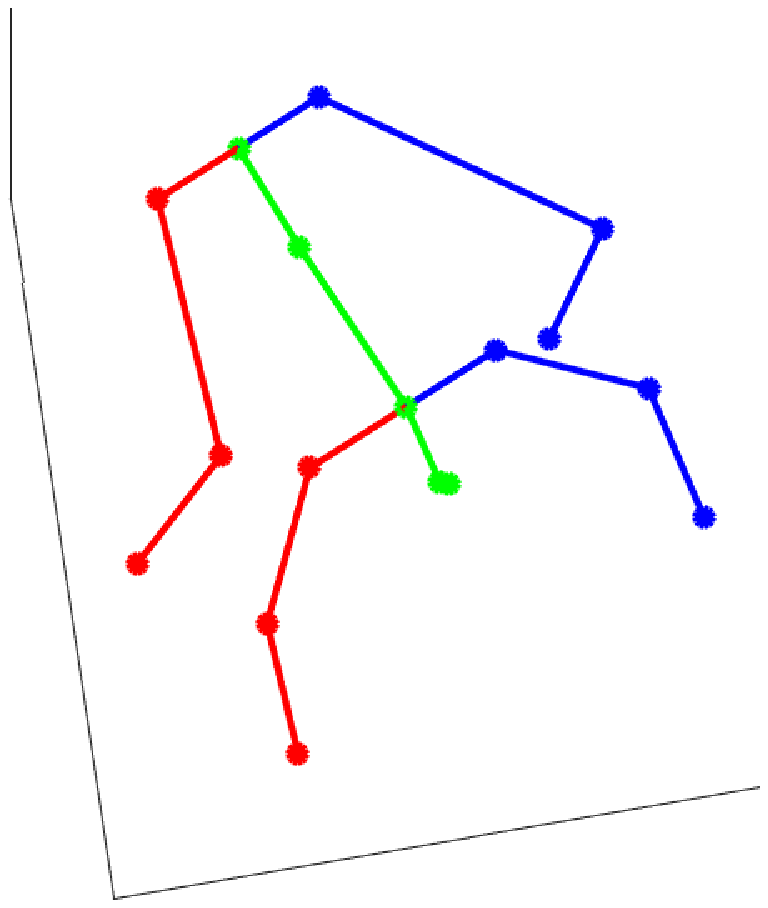}
 \end{minipage}

 \begin{minipage}{0.15\textwidth}
 \centering
     \includegraphics[width=\linewidth]{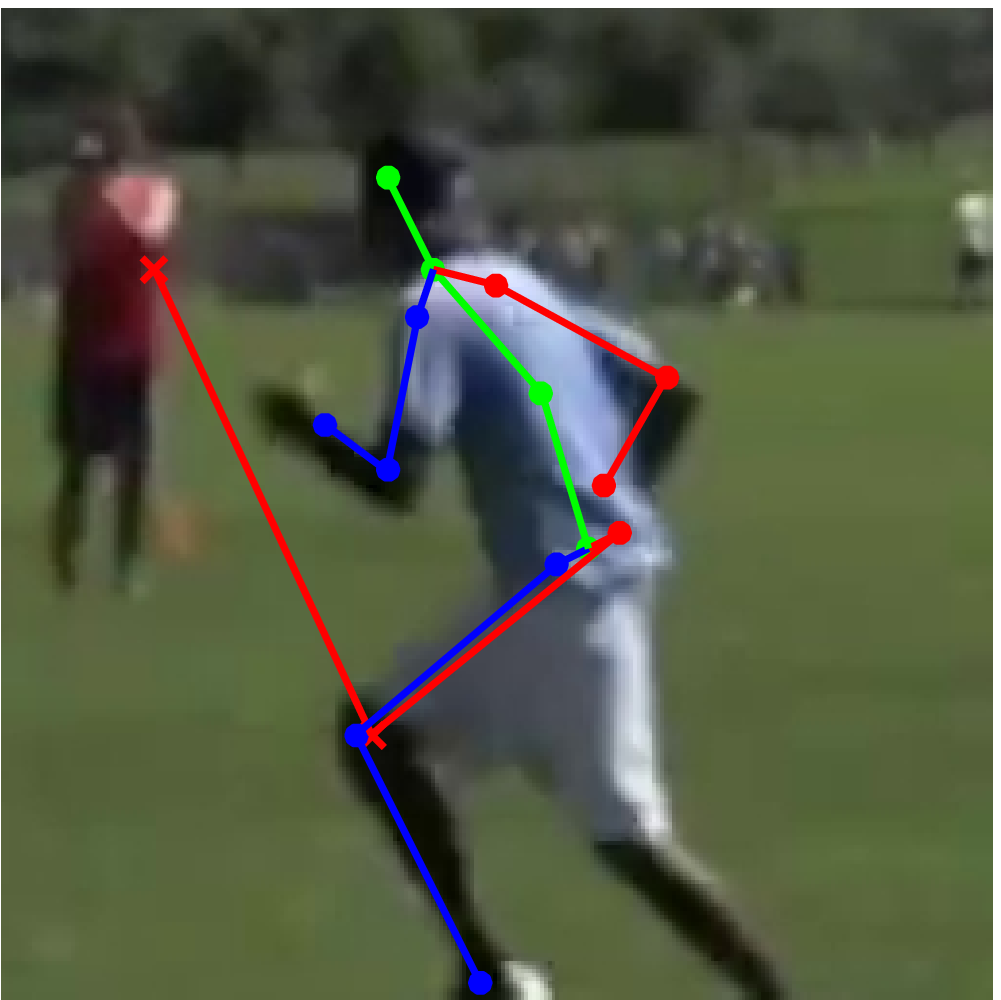}
 \end{minipage}
 \begin{minipage}{0.15\textwidth}
 \centering
     \includegraphics[width=\linewidth]{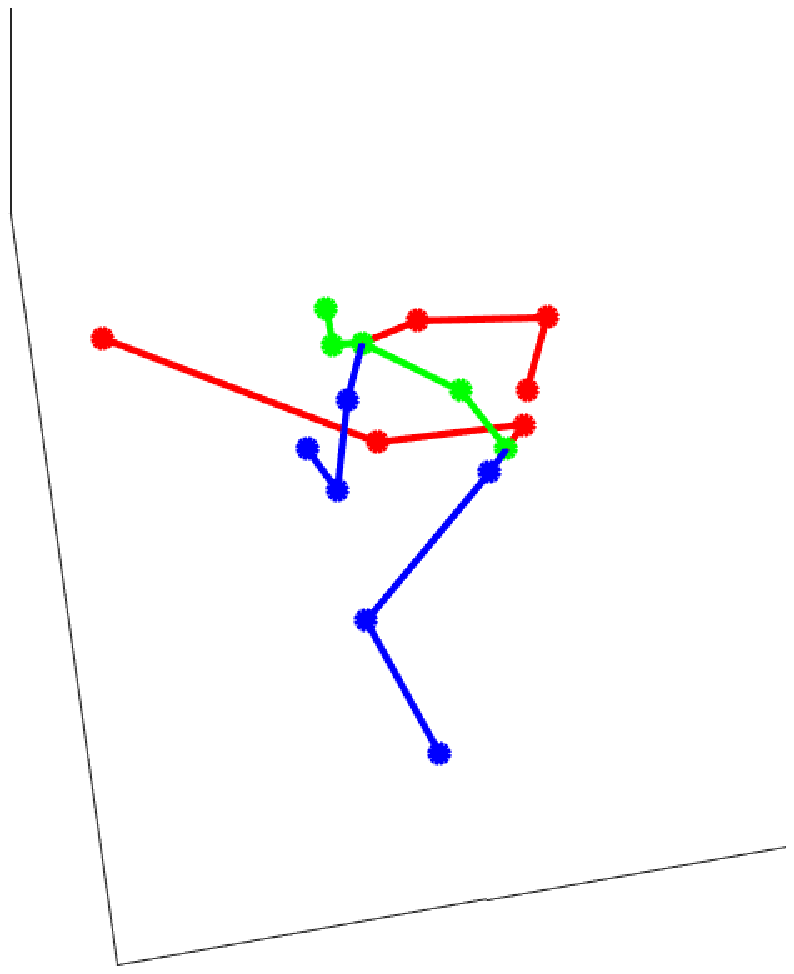}
 \end{minipage}
 \begin{minipage}{0.15\textwidth}
 \centering
     \includegraphics[width=\linewidth]{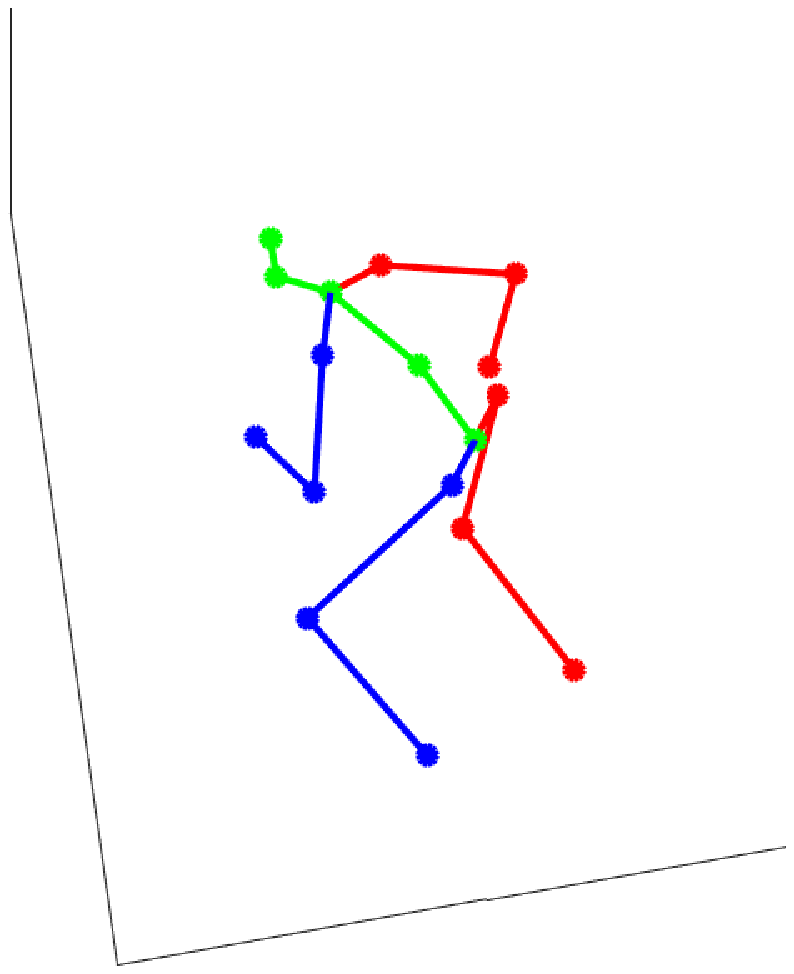}
 \end{minipage}
 \begin{minipage}{0.15\textwidth}
 \centering
     \includegraphics[width=\linewidth]{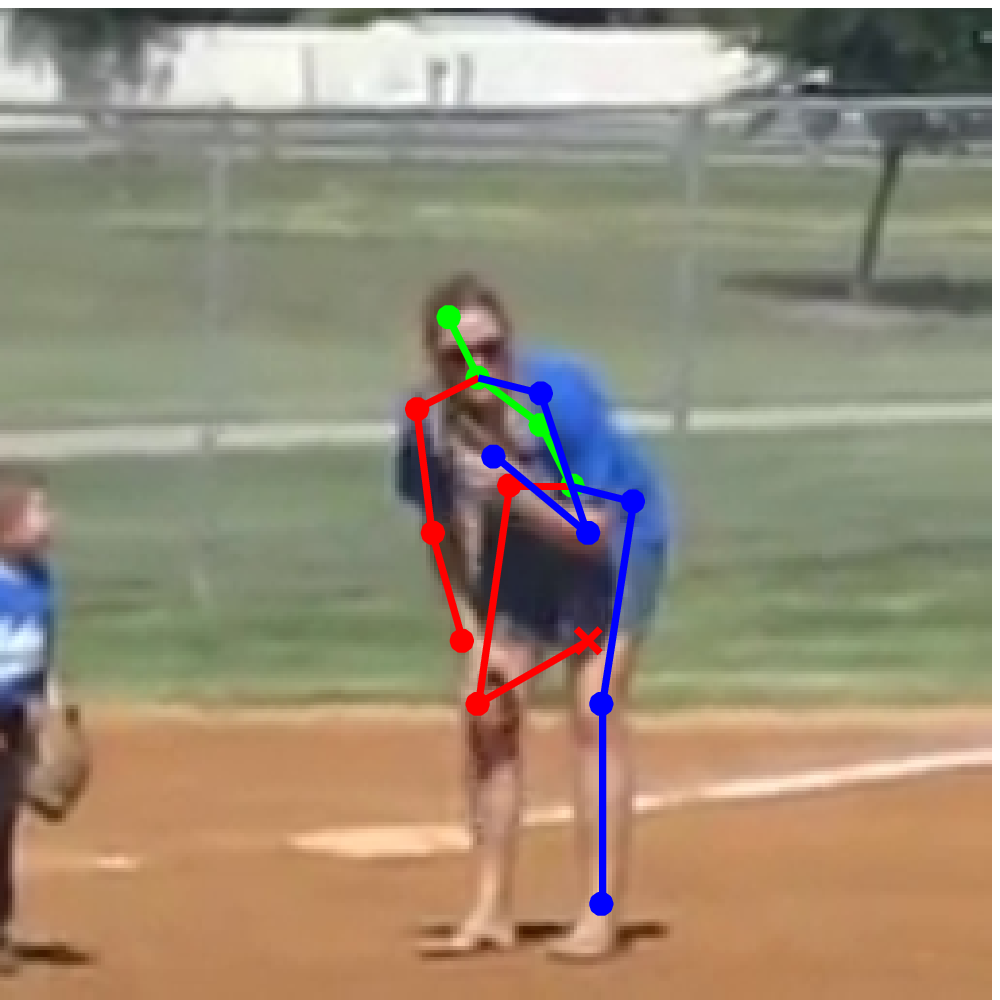}
 \end{minipage}
 \begin{minipage}{0.15\textwidth}
 \centering
     \includegraphics[width=\linewidth]{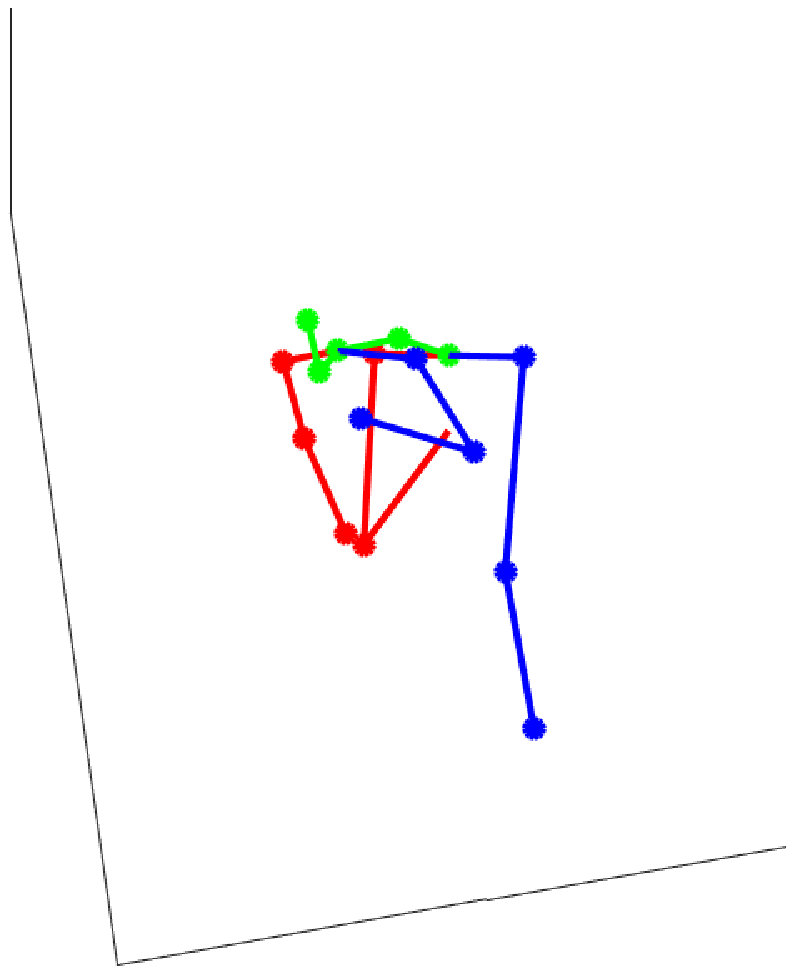}
 \end{minipage}
 \begin{minipage}{0.15\textwidth}
 \centering
     \includegraphics[width=\linewidth]{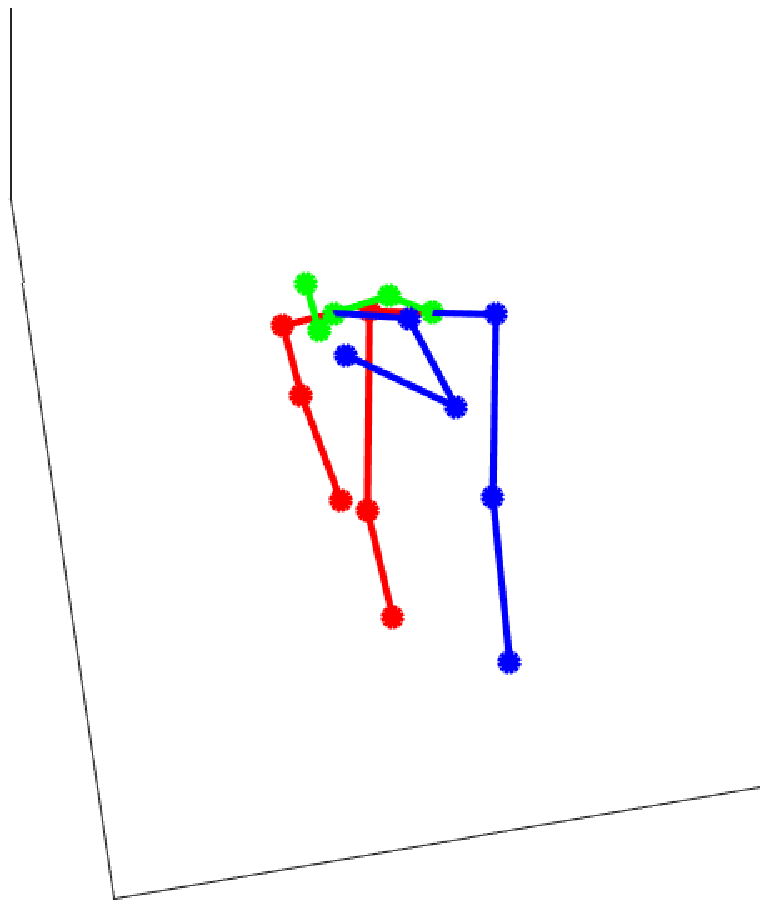}
 \end{minipage}

 \begin{minipage}{0.15\textwidth}
 \centering
     \includegraphics[width=\linewidth]{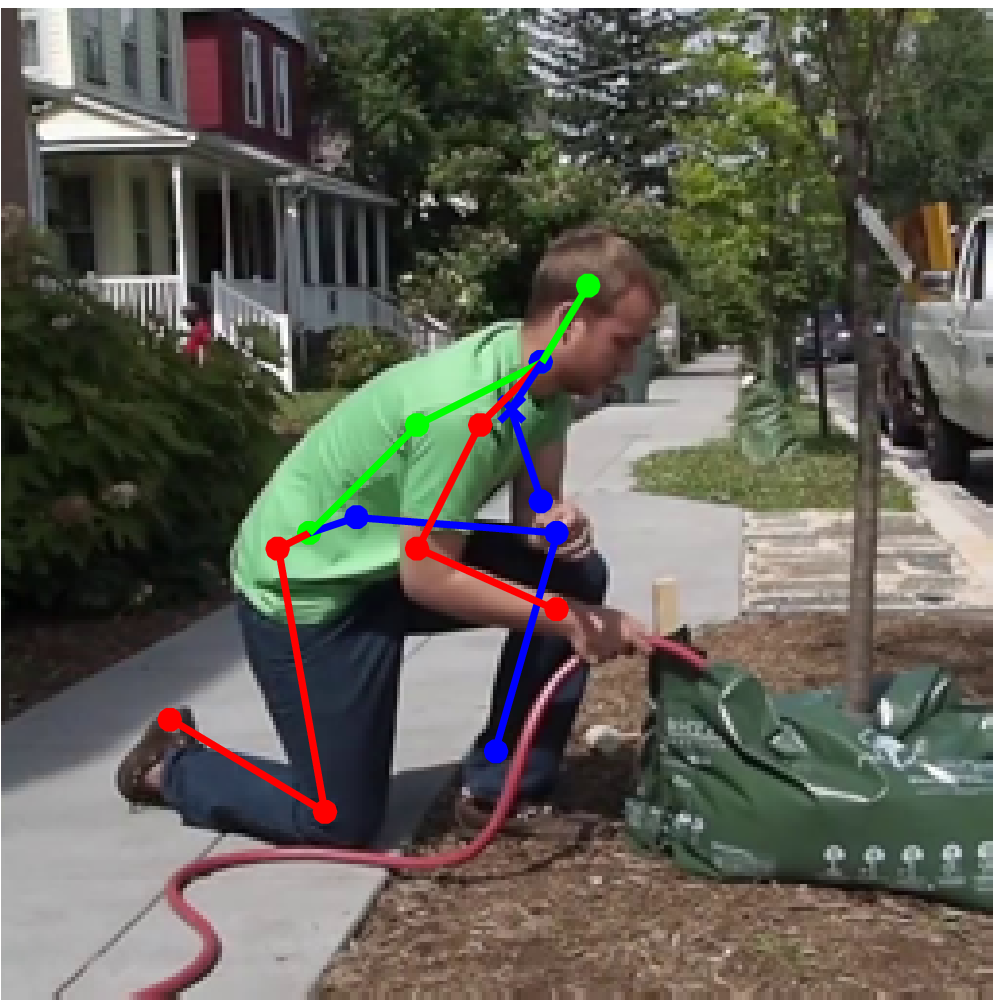}
 \end{minipage}
 \begin{minipage}{0.15\textwidth}
 \centering
     \includegraphics[width=\linewidth]{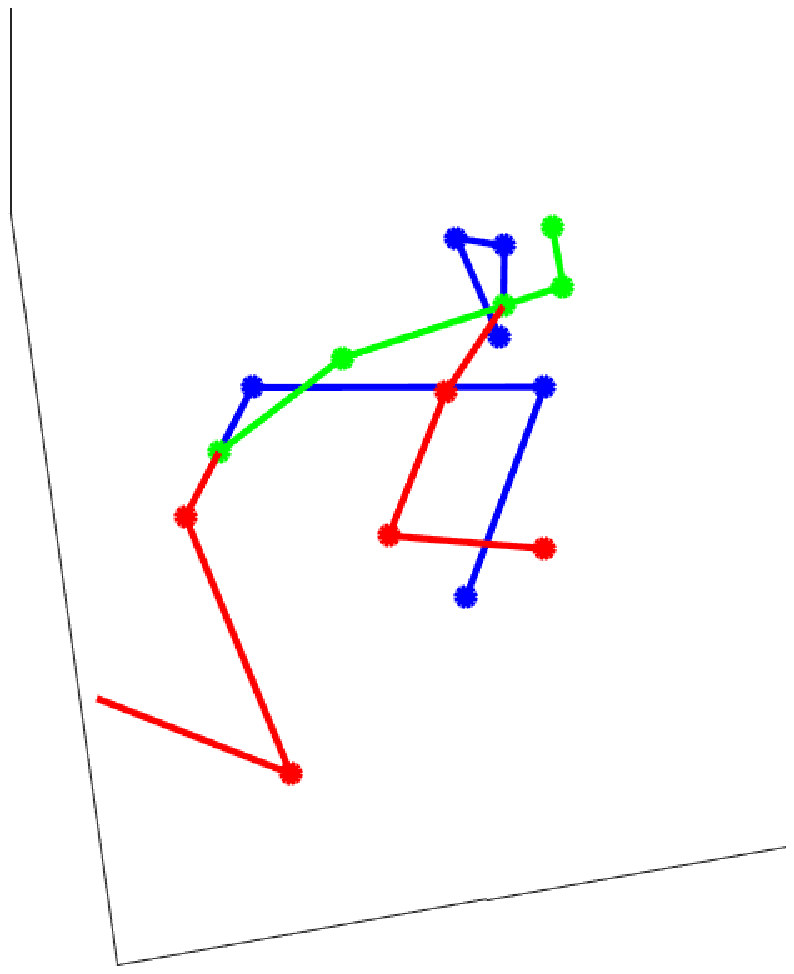}
 \end{minipage}
 \begin{minipage}{0.15\textwidth}
 \centering
     \includegraphics[width=\linewidth]{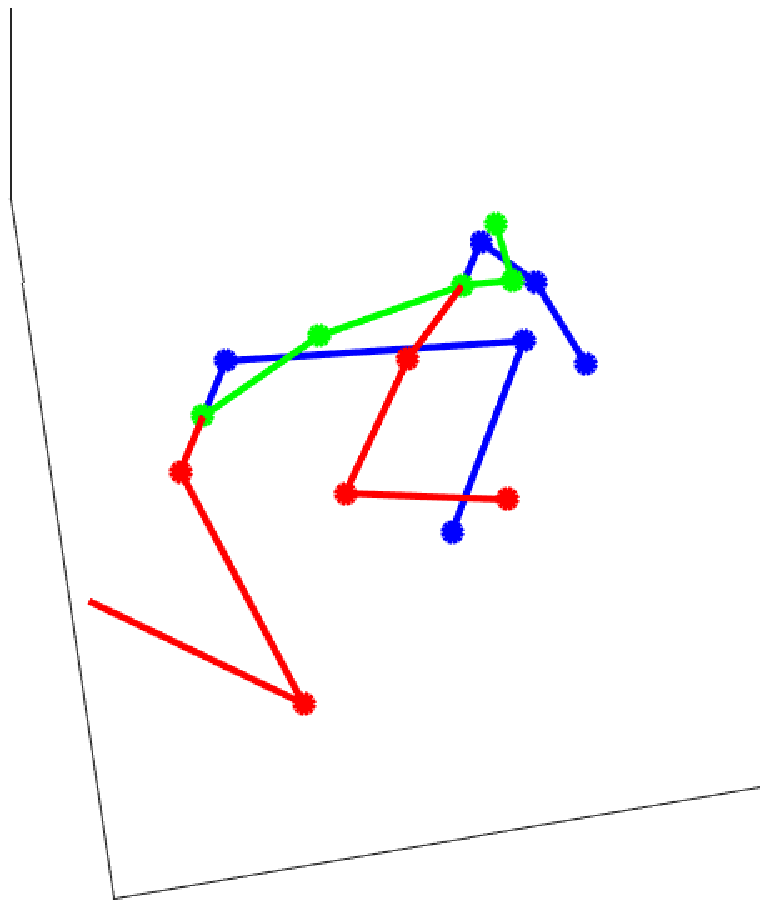}
 \end{minipage}
 \begin{minipage}{0.15\textwidth}
 \centering
     \includegraphics[width=\linewidth]{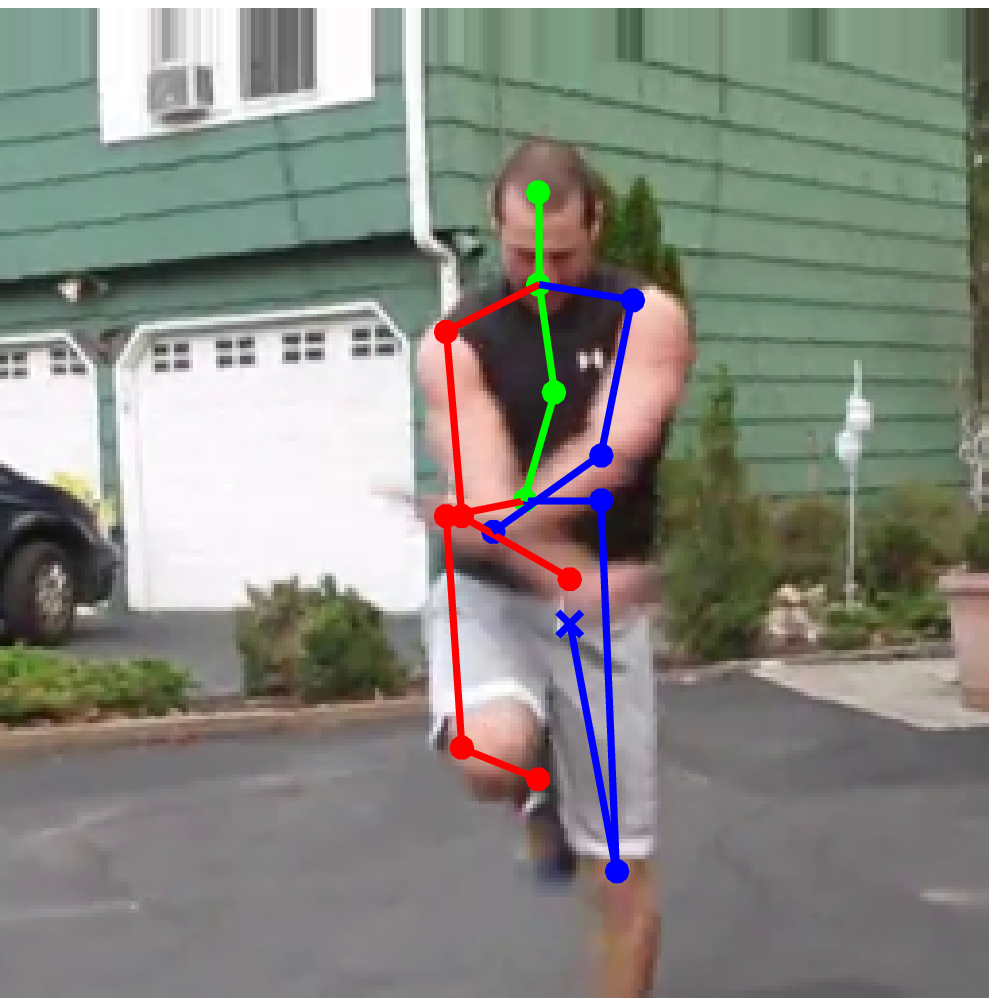}
 \end{minipage}
 \begin{minipage}{0.15\textwidth}
 \centering
     \includegraphics[width=\linewidth]{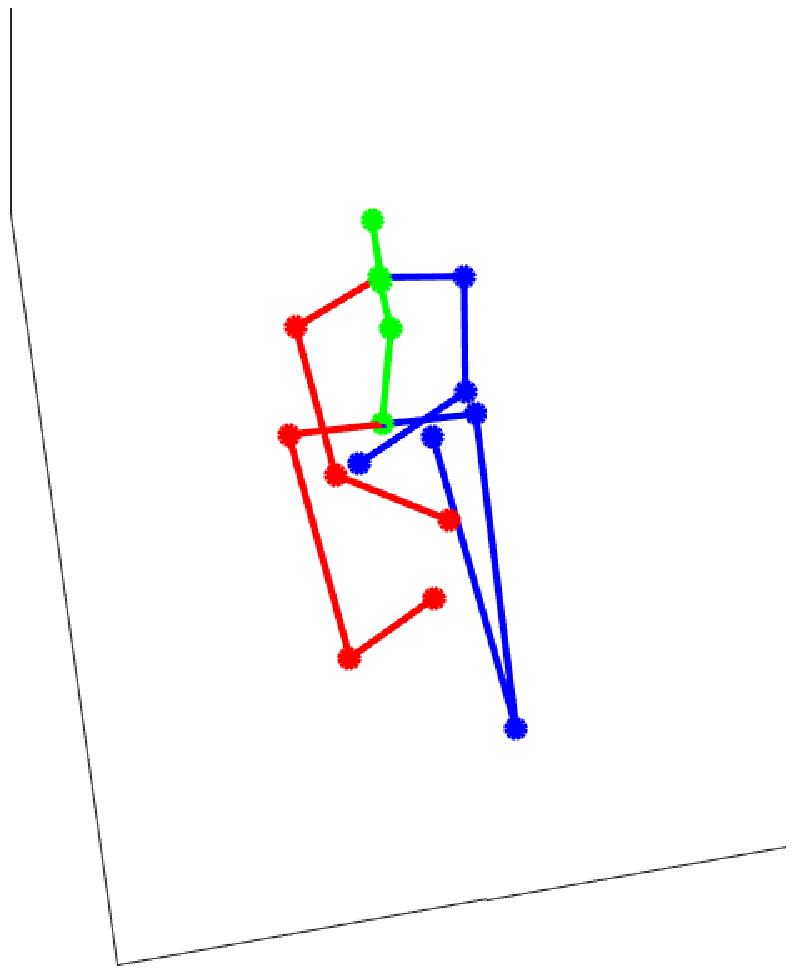}
 \end{minipage}
 \begin{minipage}{0.15\textwidth}
 \centering
     \includegraphics[width=\linewidth]{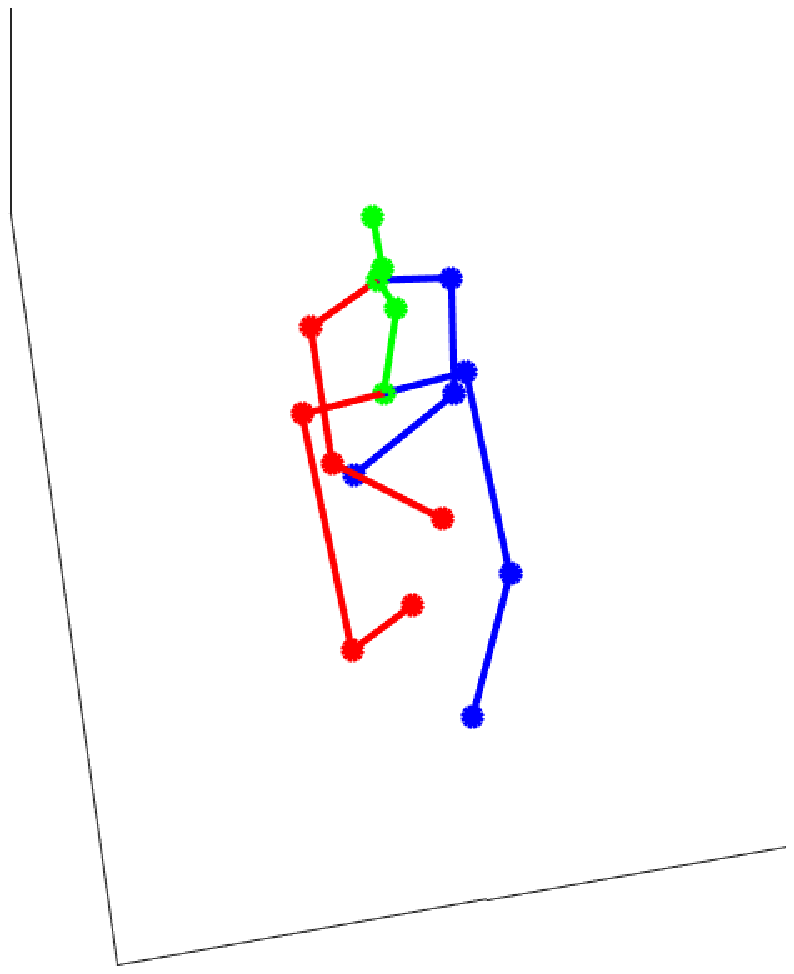}
 \end{minipage}

 \begin{minipage}{0.15\textwidth}
 \centering
     \includegraphics[width=\linewidth]{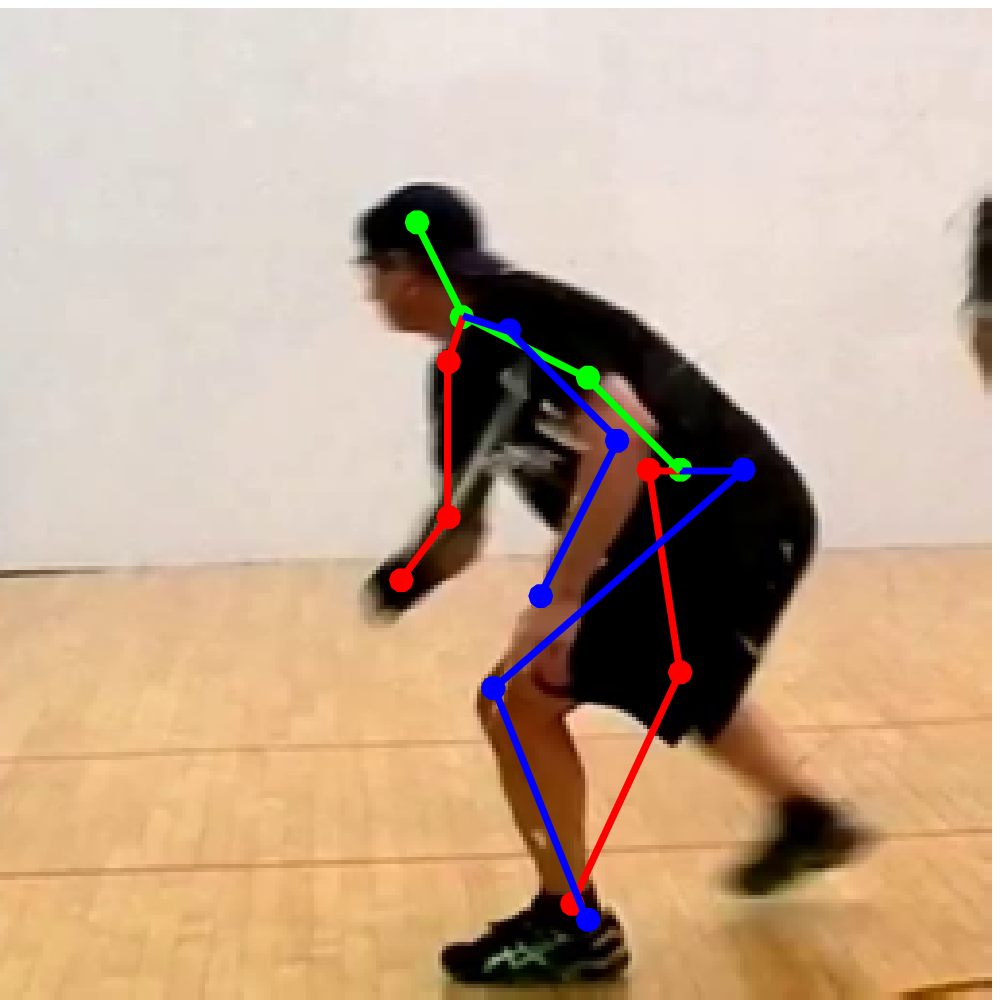}
 \end{minipage}
 \begin{minipage}{0.15\textwidth}
 \centering
     \includegraphics[width=\linewidth]{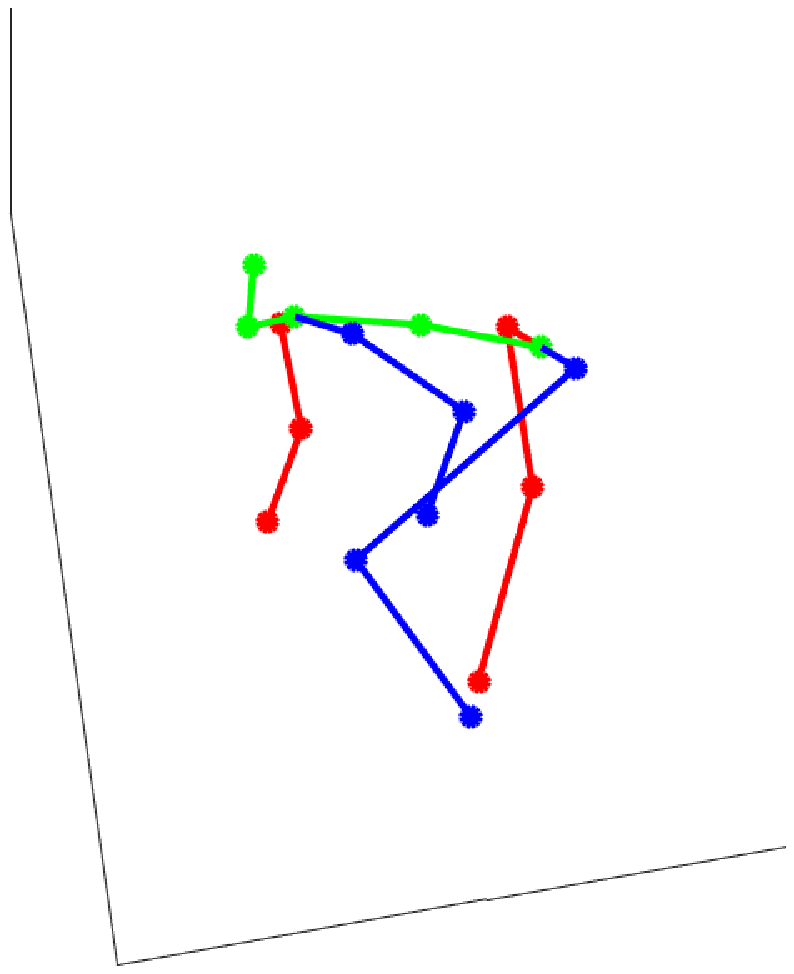}
 \end{minipage}
 \begin{minipage}{0.15\textwidth}
 \centering
     \includegraphics[width=\linewidth]{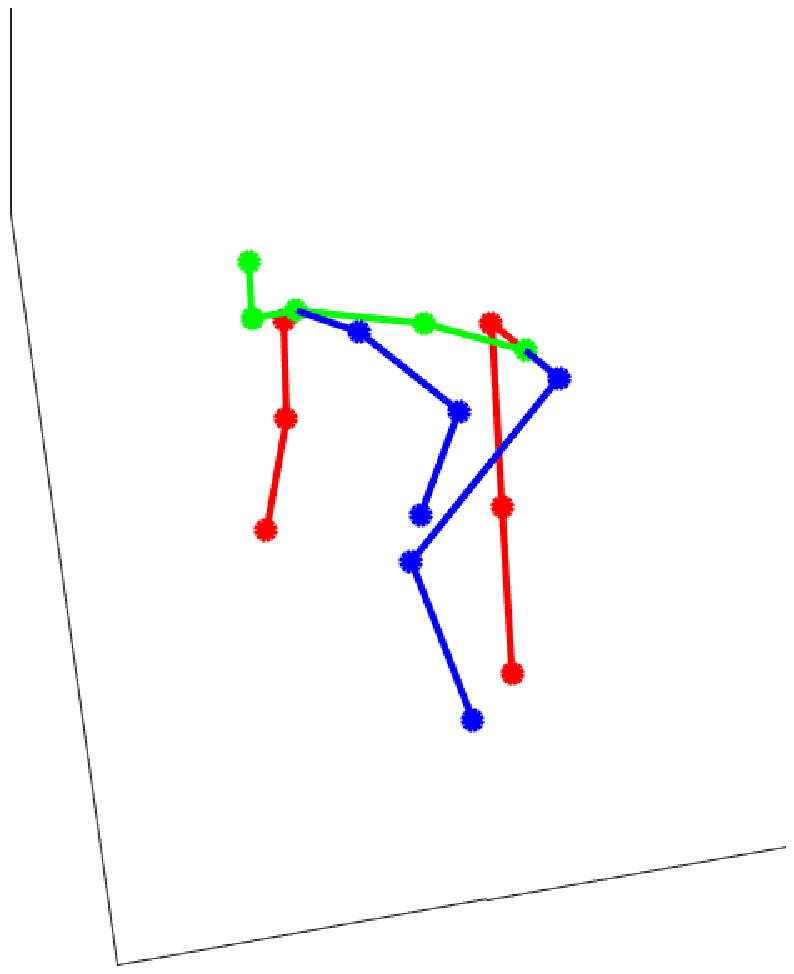}
 \end{minipage}
 \begin{minipage}{0.15\textwidth}
 \centering
     \includegraphics[width=\linewidth]{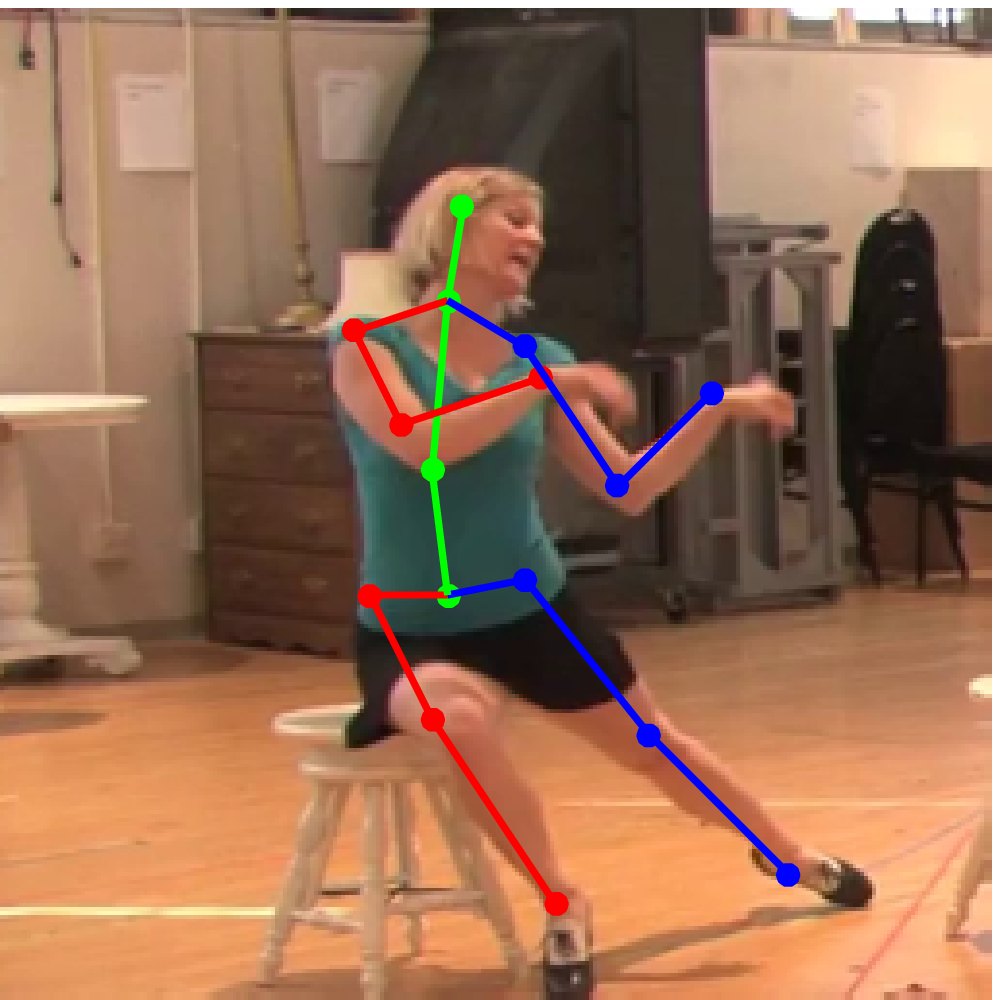}
 \end{minipage}
 \begin{minipage}{0.15\textwidth}
 \centering
     \includegraphics[width=\linewidth]{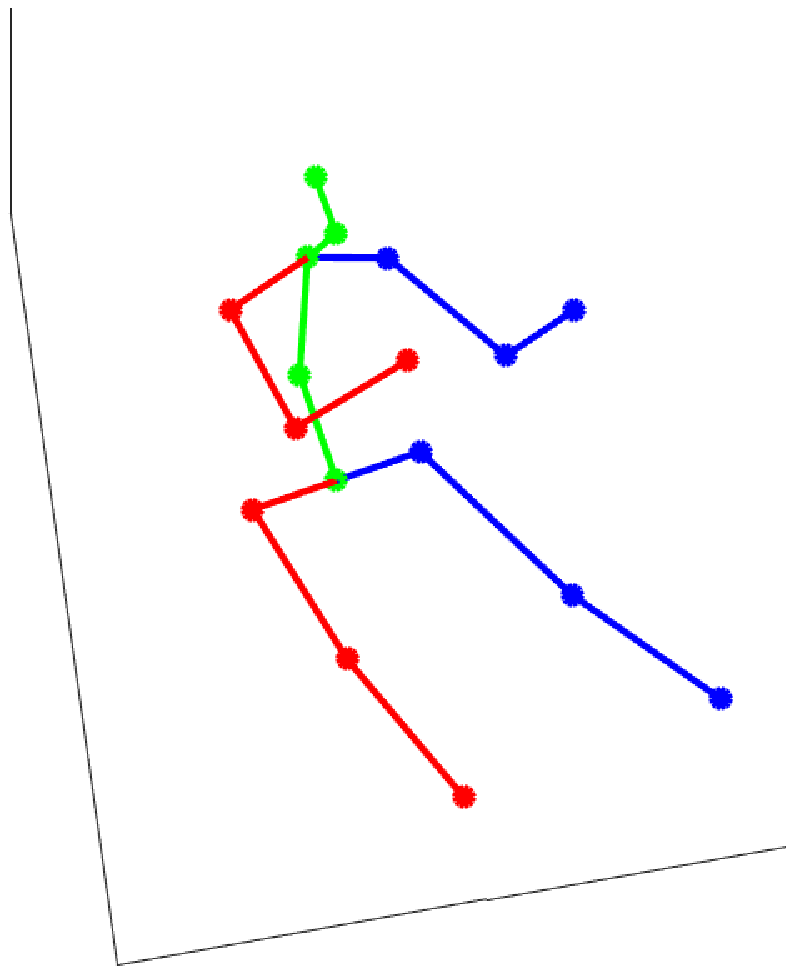}
 \end{minipage}
 \begin{minipage}{0.15\textwidth}
 \centering
     \includegraphics[width=\linewidth]{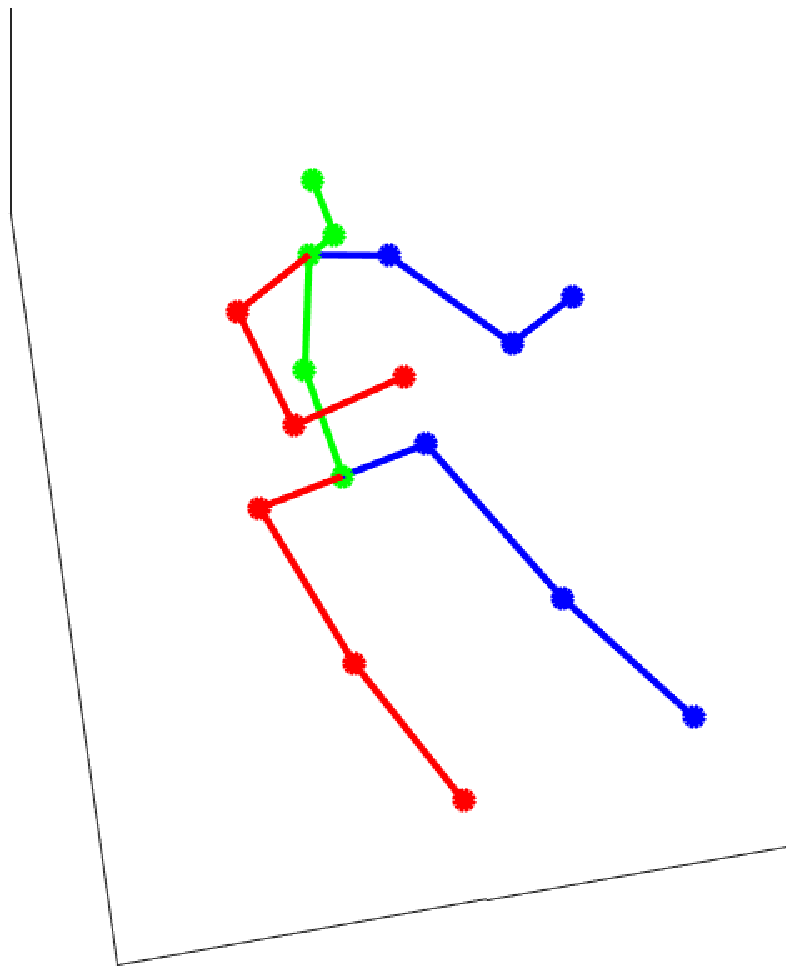}
 \end{minipage}

\caption{Qualitative results on MPII pose dataset.}
\label{fig3}
\end{figure}

Lastly, we displayed qualitative results on real world images. We used MPII human pose dataset~\cite{andriluka14cvpr} which is designed for 2D human pose estimation. 3D pose estimation results for the relational network (\textit{RN}) and the hierarchical relational network with relational dropouts (\textit{RN-hier-drop}) are provided in Figure~\ref{fig3}. We first generate 2D pose results for the images and the joints whose maximum heatmap value is less than 0.4 are treated as missing joints for \textit{RN-hier-drop}. As it can be seen in the second and third \nj{rows} of Figure~\ref{fig2}, \textit{RN-hier-drop} generates more plausible poses \nj{than} \textit{RN} when some 2D joints are wrongly detected. The last row shows failure cases which \nj{contain} noisy 2D inputs or \nj{an} unfamiliar 3D pose that is not provided during the training.

\section{Conclusion}
\label{sec:concl}
In this paper, we propose a novel method for 3D human pose estimation. The relational network designed for 3D pose estimation showed state-of-the-art performance despite its simple structure. We also proposed the relational dropout which is fitted for the relational network. The relational dropout successfully impose the robustness to the missing points while maintaining the performance of the original network. The proposed network \nj{is flexible in that it allows lots of  variations} 
in terms of its structure, group organization, and the policy of the relational dropout. The relational dropout can also \nj{be} applied to \nj{other tasks} that use relational networks.

\section*{Acknowledgments}
This work was supported by Next-Generation Information Computing Development Program through the National Research Foundation of Korea (2017M3C4A7077582).

\bibliography{egbib}

\begin{thebibliography}{40}
\providecommand{\natexlab}[1]{#1}
\providecommand{\url}[1]{\texttt{#1}}
\expandafter\ifx\csname urlstyle\endcsname\relax
  \providecommand{\doi}[1]{doi: #1}\else
  \providecommand{\doi}{doi: \begingroup \urlstyle{rm}\Url}\fi

\bibitem[Agarwal and Triggs(2006)]{agarwal2006recovering}
Ankur Agarwal and Bill Triggs.
\newblock Recovering 3d human pose from monocular images.
\newblock \emph{IEEE transactions on pattern analysis and machine
  intelligence}, 28\penalty0 (1):\penalty0 44--58, 2006.

\bibitem[Andriluka et~al.(2014)Andriluka, Pishchulin, Gehler, and
  Schiele]{andriluka14cvpr}
Mykhaylo Andriluka, Leonid Pishchulin, Peter Gehler, and Bernt Schiele.
\newblock 2d human pose estimation: New benchmark and state of the art
  analysis.
\newblock In \emph{IEEE Conference on Computer Vision and Pattern Recognition
  (CVPR)}, June 2014.

\bibitem[Bogo et~al.(2016)Bogo, Kanazawa, Lassner, Gehler, Romero, and
  Black]{bogo2016keep}
Federica Bogo, Angjoo Kanazawa, Christoph Lassner, Peter Gehler, Javier Romero,
  and Michael~J Black.
\newblock Keep it smpl: Automatic estimation of 3d human pose and shape from a
  single image.
\newblock In \emph{European Conference on Computer Vision}, pages 561--578.
  Springer, 2016.

\bibitem[Cha et~al.(2018)Cha, Lee, Cho, and Oh]{cha2018pose}
Geonho Cha, Minsik Lee, Jungchan Cho, and Songhwai Oh.
\newblock Deep pose consensus networks.
\newblock \emph{arXiv preprint arXiv:1803.08190}, 2018.

\bibitem[Chen and Ramanan(2017)]{Chen_2017_CVPR}
Ching-Hang Chen and Deva Ramanan.
\newblock 3d human pose estimation = 2d pose estimation + matching.
\newblock In \emph{The IEEE Conference on Computer Vision and Pattern
  Recognition (CVPR)}, July 2017.

\bibitem[Fang et~al.(2017)Fang, Xu, Wang, Liu, and Zhu]{fang2017learning}
Haoshu Fang, Yuanlu Xu, Wenguan Wang, Xiaobai Liu, and Song-Chun Zhu.
\newblock Learning knowledge-guided pose grammar machine for 3d human pose
  estimation.
\newblock \emph{arXiv preprint arXiv:1710.06513}, 2017.

\bibitem[Grinciunaite et~al.(2016)Grinciunaite, Gudi, Tasli, and den
  Uyl]{grinciunaite2016human}
Agne Grinciunaite, Amogh Gudi, Emrah Tasli, and Marten den Uyl.
\newblock Human pose estimation in space and time using 3d cnn.
\newblock In \emph{European Conference on Computer Vision}, pages 32--39.
  Springer, 2016.

\bibitem[He et~al.(2016{\natexlab{a}})He, Zhang, Ren, and Sun]{he2016deep}
Kaiming He, Xiangyu Zhang, Shaoqing Ren, and Jian Sun.
\newblock Deep residual learning for image recognition.
\newblock In \emph{Proceedings of the IEEE conference on computer vision and
  pattern recognition}, pages 770--778, 2016{\natexlab{a}}.

\bibitem[He et~al.(2016{\natexlab{b}})He, Zhang, Ren, and Sun]{he2016identity}
Kaiming He, Xiangyu Zhang, Shaoqing Ren, and Jian Sun.
\newblock Identity mappings in deep residual networks.
\newblock In \emph{European Conference on Computer Vision}, pages 630--645.
  Springer, 2016{\natexlab{b}}.

\bibitem[Hossain and Little(2017)]{hossain2017exploiting}
Mir Rayat~Imtiaz Hossain and James~J Little.
\newblock Exploiting temporal information for 3d pose estimation.
\newblock \emph{arXiv preprint arXiv:1711.08585}, 2017.

\bibitem[Ioffe and Szegedy(2015)]{ioffe15}
Sergey Ioffe and Christian Szegedy.
\newblock Batch normalization: Accelerating deep network training by reducing
  internal covariate shift.
\newblock In Francis Bach and David Blei, editors, \emph{Proceedings of the
  32nd International Conference on Machine Learning}, volume~37 of
  \emph{Proceedings of Machine Learning Research}, pages 448--456, Lille,
  France, 07--09 Jul 2015. PMLR.

\bibitem[Ionescu et~al.(2011)Ionescu, Li, and Sminchisescu]{ionescu2011latent}
Catalin Ionescu, Fuxin Li, and Cristian Sminchisescu.
\newblock Latent structured models for human pose estimation.
\newblock In \emph{Computer Vision (ICCV), 2011 IEEE International Conference
  on}, pages 2220--2227. IEEE, 2011.

\bibitem[Ionescu et~al.(2014)Ionescu, Papava, Olaru, and
  Sminchisescu]{ionescu2014human3}
Catalin Ionescu, Dragos Papava, Vlad Olaru, and Cristian Sminchisescu.
\newblock Human3. 6m: Large scale datasets and predictive methods for 3d human
  sensing in natural environments.
\newblock \emph{IEEE transactions on pattern analysis and machine
  intelligence}, 36\penalty0 (7):\penalty0 1325--1339, 2014.

\bibitem[Kingma and Ba(2014)]{kingma2014adam}
Diederik~P Kingma and Jimmy Ba.
\newblock Adam: A method for stochastic optimization.
\newblock \emph{arXiv preprint arXiv:1412.6980}, 2014.

\bibitem[Lassner et~al.(2017)Lassner, Romero, Kiefel, Bogo, Black, and
  Gehler]{lassner2017unite}
Christoph Lassner, Javier Romero, Martin Kiefel, Federica Bogo, Michael~J
  Black, and Peter~V Gehler.
\newblock Unite the people: Closing the loop between 3d and 2d human
  representations.
\newblock In \emph{IEEE Conf. on Computer Vision and Pattern Recognition
  (CVPR)}, 2017.

\bibitem[Lee et~al.(2016)Lee, Cho, and Oh]{lee2016consensus}
Minsik Lee, Jungchan Cho, and Songhwai Oh.
\newblock Consensus of non-rigid reconstructions.
\newblock In \emph{Proceedings of the IEEE Conference on Computer Vision and
  Pattern Recognition}, pages 4670--4678, 2016.

\bibitem[Li and Chan(2014)]{li20143d}
Sijin Li and Antoni~B Chan.
\newblock 3d human pose estimation from monocular images with deep
  convolutional neural network.
\newblock In \emph{Asian Conference on Computer Vision}, pages 332--347.
  Springer, 2014.

\bibitem[Li et~al.(2015)Li, Zhang, and Chan]{li2015maximum}
Sijin Li, Weichen Zhang, and Antoni~B Chan.
\newblock Maximum-margin structured learning with deep networks for 3d human
  pose estimation.
\newblock In \emph{Proceedings of the IEEE International Conference on Computer
  Vision}, pages 2848--2856, 2015.

\bibitem[Martinez et~al.(2017)Martinez, Hossain, Romero, and
  Little]{Martinez_2017_ICCV}
Julieta Martinez, Rayat Hossain, Javier Romero, and James~J. Little.
\newblock A simple yet effective baseline for 3d human pose estimation.
\newblock In \emph{The IEEE International Conference on Computer Vision
  (ICCV)}, Oct 2017.

\bibitem[Mehta et~al.(2017)Mehta, Sridhar, Sotnychenko, Rhodin, Shafiei,
  Seidel, Xu, Casas, and Theobalt]{mehta2017vnect}
Dushyant Mehta, Srinath Sridhar, Oleksandr Sotnychenko, Helge Rhodin, Mohammad
  Shafiei, Hans-Peter Seidel, Weipeng Xu, Dan Casas, and Christian Theobalt.
\newblock Vnect: Real-time 3d human pose estimation with a single rgb camera.
\newblock \emph{ACM Transactions on Graphics (TOG)}, 36\penalty0 (4):\penalty0
  44, 2017.

\bibitem[Moreno-Noguer(2017)]{moreno20173d}
Francesc Moreno-Noguer.
\newblock 3d human pose estimation from a single image via distance matrix
  regression.
\newblock In \emph{2017 IEEE Conference on Computer Vision and Pattern
  Recognition (CVPR)}, pages 1561--1570. IEEE, 2017.

\bibitem[Newell et~al.(2016)Newell, Yang, and Deng]{newell2016stacked}
Alejandro Newell, Kaiyu Yang, and Jia Deng.
\newblock Stacked hourglass networks for human pose estimation.
\newblock In \emph{European Conference on Computer Vision}, pages 483--499.
  Springer, 2016.

\bibitem[Onishi et~al.(2008)Onishi, Takiguchi, and Ariki]{onishi20083d}
Katsunori Onishi, Tetsuya Takiguchi, and Yasuo Ariki.
\newblock 3d human posture estimation using the hog features from monocular
  image.
\newblock In \emph{Pattern Recognition, 2008. ICPR 2008. 19th International
  Conference on}, pages 1--4. IEEE, 2008.

\bibitem[Park et~al.(2016)Park, Hwang, and Kwak]{park_eccvw_2016}
Sungheon Park, Jihye Hwang, and Nojun Kwak.
\newblock 3d human pose estimation using convolutional neural networks with 2d
  pose information.
\newblock In Gang Hua and Herv{\'e} J{\'e}gou, editors, \emph{Computer Vision
  -- ECCV 2016 Workshops}, pages 156--169, Cham, 2016. Springer International
  Publishing.

\bibitem[Pavlakos et~al.(2017{\natexlab{a}})Pavlakos, Zhou, Derpanis, and
  Daniilidis]{Pavlakos_2017_CVPR}
Georgios Pavlakos, Xiaowei Zhou, Konstantinos~G. Derpanis, and Kostas
  Daniilidis.
\newblock Harvesting multiple views for marker-less 3d human pose annotations.
\newblock In \emph{The IEEE Conference on Computer Vision and Pattern
  Recognition (CVPR)}, July 2017{\natexlab{a}}.

\bibitem[Pavlakos et~al.(2017{\natexlab{b}})Pavlakos, Zhou, Derpanis, and
  Daniilidis]{pavlakos2017coarse}
Georgios Pavlakos, Xiaowei Zhou, Konstantinos~G Derpanis, and Kostas
  Daniilidis.
\newblock Coarse-to-fine volumetric prediction for single-image 3d human pose.
\newblock In \emph{Computer Vision and Pattern Recognition (CVPR), 2017 IEEE
  Conference on}, pages 1263--1272. IEEE, 2017{\natexlab{b}}.

\bibitem[Rogez et~al.(2008)Rogez, Rihan, Ramalingam, Orrite, and
  Torr]{rogez2008randomized}
Gr{\'e}gory Rogez, Jonathan Rihan, Srikumar Ramalingam, Carlos Orrite, and
  Philip~HS Torr.
\newblock Randomized trees for human pose detection.
\newblock In \emph{Computer Vision and Pattern Recognition, 2008. CVPR 2008.
  IEEE Conference on}, pages 1--8. IEEE, 2008.

\bibitem[Rogez et~al.(2017)Rogez, Weinzaepfel, and Schmid]{rogez2017lcr}
Gregory Rogez, Philippe Weinzaepfel, and Cordelia Schmid.
\newblock Lcr-net: Localization-classification-regression for human pose.
\newblock In \emph{CVPR 2017-IEEE Conference on Computer Vision \& Pattern
  Recognition}, 2017.

\bibitem[Santoro et~al.(2017)Santoro, Raposo, Barrett, Malinowski, Pascanu,
  Battaglia, and Lillicrap]{santoro2017}
Adam Santoro, David Raposo, David G.~T. Barrett, Mateusz Malinowski, Razvan
  Pascanu, Peter Battaglia, and Timothy~P. Lillicrap.
\newblock A simple neural network module for relational reasoning.
\newblock \emph{CoRR}, abs/1706.01427, 2017.

\bibitem[Sanzari et~al.(2016)Sanzari, Ntouskos, and Pirri]{sanzari2016bayesian}
Marta Sanzari, Valsamis Ntouskos, and Fiora Pirri.
\newblock Bayesian image based 3d pose estimation.
\newblock In \emph{European Conference on Computer Vision}, pages 566--582.
  Springer, 2016.

\bibitem[Srivastava et~al.(2014)Srivastava, Hinton, Krizhevsky, Sutskever, and
  Salakhutdinov]{srivastava2014dropout}
Nitish Srivastava, Geoffrey Hinton, Alex Krizhevsky, Ilya Sutskever, and Ruslan
  Salakhutdinov.
\newblock Dropout: A simple way to prevent neural networks from overfitting.
\newblock \emph{The Journal of Machine Learning Research}, 15\penalty0
  (1):\penalty0 1929--1958, 2014.

\bibitem[Sun et~al.(2017)Sun, Shang, Liang, and Wei]{sun2017compositional}
Xiao Sun, Jiaxiang Shang, Shuang Liang, and Yichen Wei.
\newblock Compositional human pose regression.
\newblock In \emph{The IEEE International Conference on Computer Vision
  (ICCV)}, volume~2, 2017.

\bibitem[Tekin et~al.(2016)Tekin, Katircioglu, Salzmann, Lepetit, and
  Fua]{tekin_2016_BMVC}
Bugra Tekin, Isinsu Katircioglu, Mathieu Salzmann, Vincent Lepetit, and Pascal
  Fua.
\newblock Structured prediction of 3d human pose with deep neural networks.
\newblock In \emph{Proceedings of the British Machine Vision Conference
  (BMVC)}, pages 130.1--130.11. BMVA Press, September 2016.

\bibitem[Tekin et~al.(2017)Tekin, Marquez-Neila, Salzmann, and
  Fua]{Tekin_2017_ICCV}
Bugra Tekin, Pablo Marquez-Neila, Mathieu Salzmann, and Pascal Fua.
\newblock Learning to fuse 2d and 3d image cues for monocular body pose
  estimation.
\newblock In \emph{The IEEE International Conference on Computer Vision
  (ICCV)}, Oct 2017.

\bibitem[Tome et~al.(2017)Tome, Russell, and Agapito]{tome2017lifting}
Denis Tome, Christopher Russell, and Lourdes Agapito.
\newblock Lifting from the deep: Convolutional 3d pose estimation from a single
  image.
\newblock \emph{CVPR 2017 Proceedings}, pages 2500--2509, 2017.

\bibitem[Yang et~al.(2018)Yang, Ouyang, Wang, Ren, Li, and Wang]{yang20183d}
Wei Yang, Wanli Ouyang, Xiaolong Wang, Jimmy Ren, Hongsheng Li, and Xiaogang
  Wang.
\newblock 3d human pose estimation in the wild by adversarial learning.
\newblock \emph{arXiv preprint arXiv:1803.09722}, 2018.

\bibitem[Yasin et~al.(2016)Yasin, Iqbal, Kruger, Weber, and
  Gall]{Yasin_2016_CVPR}
Hashim Yasin, Umar Iqbal, Bjorn Kruger, Andreas Weber, and Juergen Gall.
\newblock A dual-source approach for 3d pose estimation from a single image.
\newblock In \emph{The IEEE Conference on Computer Vision and Pattern
  Recognition (CVPR)}, June 2016.

\bibitem[Zhou et~al.(2016)Zhou, Zhu, Leonardos, Derpanis, and
  Daniilidis]{zhou2016sparseness}
Xiaowei Zhou, Menglong Zhu, Spyridon Leonardos, Konstantinos~G Derpanis, and
  Kostas Daniilidis.
\newblock Sparseness meets deepness: 3d human pose estimation from monocular
  video.
\newblock In \emph{Proceedings of the IEEE conference on computer vision and
  pattern recognition}, pages 4966--4975, 2016.

\bibitem[Zhou et~al.(2018)Zhou, Zhu, Pavlakos, Leonardos, Derpanis, and
  Daniilidis]{zhou2018monocap}
Xiaowei Zhou, Menglong Zhu, Georgios Pavlakos, Spyridon Leonardos,
  Konstantinos~G Derpanis, and Kostas Daniilidis.
\newblock Monocap: Monocular human motion capture using a cnn coupled with a
  geometric prior.
\newblock \emph{IEEE Transactions on Pattern Analysis and Machine
  Intelligence}, 2018.

\bibitem[Zhou et~al.(2017)Zhou, Huang, Sun, Xue, and Wei]{zhou2017towards}
Xingyi Zhou, Qixing Huang, Xiao Sun, Xiangyang Xue, and Yichen Wei.
\newblock Towards 3d human pose estimation in the wild: a weakly-supervised
  approach.
\newblock In \emph{IEEE International Conference on Computer Vision}, 2017.

\end{thebibliography}
\end{document}